\documentclass[acmtog,screen]{acmart}
\acmSubmissionID{211}

\usepackage{graphicx}
\usepackage[capitalize]{cleveref}

\usepackage{orcidlink}

\usepackage{wrapfig}
\usepackage{multirow}
\usepackage{tabularx}
\usepackage{algorithmic}
\usepackage[ruled]{algorithm2e}

\SetAlFnt{\small}
\SetAlCapFnt{\small}
\SetAlCapNameFnt{\small}
\SetAlCapHSkip{0pt}

\usepackage{mathtools}
\usepackage{balance}

\usepackage{float}
\usepackage{enumitem}
\usepackage{comment}
\usepackage{pifont}
\usepackage[toc,page]{appendix}

\usepackage{soul}
\usepackage{acronym}
\usepackage{fancybox}
\usepackage{epigraph}
\usepackage{dirtytalk}
\usepackage{bm}
\usepackage{fontawesome}
\usepackage{colortbl}
\usepackage{booktabs}

\usepackage{xr}
\usepackage{tikz}
\usepackage{microtype}
\usepackage{transparent}
\usepackage{subfig}
\usepackage{hyphenat}
\usepackage[toc,page]{appendix}

\definecolor{citecolor}{HTML}{0071bc}
\definecolor{frontcolor}{HTML}{325ea5}
\definecolor{backcolor}{HTML}{a58b77}
\definecolor{sidecolor}{HTML}{10768c}
\definecolor{skincolor}{HTML}{dcb7b7}
\definecolor{darkred}{rgb}{0.6, 0.1, 0.05}
\definecolor{DeltaColor}{rgb}{0.039,0.73,0.71}
\definecolor{SigmaColor}{rgb}{0.98,0.45,0.0}
\definecolor{AlphaColor}{rgb}{0,0,0.8}
\definecolor{BetaColor}{rgb}{0.8,0,0.8}
\definecolor{GammaColor}{rgb}{0.514,0.34,0.224}
\definecolor{EpsilonColor}{rgb}{0.353,0.725,0.906}
\definecolor{PurpleColor}{HTML}{9839ff}
\definecolor{BadColor}{HTML}{C0392B}
\definecolor{OrangeColor}{rgb}{0.914,0.541,0.0.141}
\definecolor{GreenColor}{HTML}{00ab41}
\definecolor{LightBlue}{HTML}{7dbaf3}
\definecolor{RedColor}{rgb}{0.949,0.275, 0.224}
\definecolor{LightCyan}{rgb}{0.88,1,1}
\definecolor{Gray}{gray}{0.85}
\definecolor{LightGray}{gray}{0.70}

\definecolor{greenprior}{HTML}{34a853}
\definecolor{redprior}{HTML}{ea4335}
\definecolor{blueprior}{HTML}{4285f4}

\definecolor{bestcolor}{rgb}{1, 0.5, 0.25}
\definecolor{secondbestcolor}{rgb}{1, 0.8, 0.5}

\newcommand{\bgood}[1]{\textcolor{GreenColor}{#1\%}}
\newcommand{\bbad}[1]{\textcolor{RedColor}{#1\%}}
\newcommand{\btie}[1]{\textcolor{OrangeColor}{#1\%}}

\newcommand{\etc}{\mbox{etc}\xspace}
\newcommand{\etal}{\mbox{et al.}\xspace}
\newcommand{\ie}{\mbox{i.e.}\xspace}
\newcommand{\eg}{\mbox{e.g.}\xspace}

\newcolumntype{a}{>{\columncolor{Gray}}c}

\newcommand{\camera}[1]{{\color{black} #1}}

\newcommand{\qheading}[1]{\noindent\textbf{#1.}}

\newcommand{\zheading}[1]{\textbf{#1.}}
\newcommand{\lzheading}[1]{\medskip\textbf{#1.}}
\newcommand{\qaheading}[1]{\medskip\noindent\textbf{#1?}}

\newcommand{\CHECK}[1]{\textcolor{black}{#1}\xspace}

\newcommand{\UPDATE}[1]{\textcolor{black}{#1}\xspace}

\makeatletter
\newcommand*{\addFileDependency}[1]{%
  \typeout{(#1)}
  \@addtofilelist{#1}
  \IfFileExists{#1}{}{\typeout{No file #1.}}
}
\makeatother

\newlength\savewidth\newcommand\shline{\noalign{\global\savewidth\arrayrulewidth
  \global\arrayrulewidth 1pt}\hline\noalign{\global\arrayrulewidth\savewidth}}

\newcommand{\vid}{\href{https://youtu.be/0hpXH2tVPk4}{\textcolor{magenta}{\tt\textit{video}}\xspace}}
\newcommand{\page}{\href{https://puzzleavatar.is.tue.mpg.de/}{\textcolor{magenta}{\xspace\tt\textit{puzzleavatar.is.tue.mpg.de}}\xspace}}
\newcommand{\specific}[1]{\texttt{\small{#1}}\xspace}

\newcommand{\xmark}{\textcolor{RedColor}{\ding{55}}\xspace}
\newcommand{\cmark}{\textcolor{GreenColor}{\ding{51}}\xspace}

\newcommand{\wrt}{\mbox{w.r.t.}\xspace}

\newcommand{\dmtet}{\mbox{DMTet}\xspace}
\newcommand{\smplx}{\mbox{SMPL-X}\xspace}

\newcommand{\mvd}{\mbox{MVDream}\xspace}

\newcommand{\sds}{Score Distillation Sampling (SDS)\xspace}

\newcommand{\sota}{state-of-the-art\xspace}

\newcommand{\subsec}{PuzzleBooth}
\newcommand{\modelname}{\mbox{PuzzleAvatar}\xspace}
\newcommand{\modelnameLong}{Assembling 3D Avatars from Personal Albums}
\newcommand{\ourtitle}{\modelname: \modelnameLong}
\newcommand{\dataname}{\textcolor{black}{\mbox{PuzzleIOI}}\xspace}

\acrodef{amt}[AMT]{Amazon Mechanical Turk}

\begin{document}

\title{\ourtitle}

\author{Yuliang Xiu}
\email{yuliang.xiu@tuebingen.mpg.de}
\orcid{0000-0003-0165-5909}
\affiliation{%
  \institution{Max Planck Institute for Intelligent Systems}
  \country{Germany}
}

\author{Yufei Ye}
\email{yeyf13.judy@gmail.com}
\orcid{0000-0001-8767-0848}
\affiliation{%
  \institution{Max Planck Institute for Intelligent Systems}
  \country{Germany}
}
\affiliation{%
  \institution{Carnegie Mellon University}
  \country{USA}
}

\author{Zhen Liu}
\email{zhen.liu@tuebingen.mpg.de}
\orcid{0009-0009-4599-3206}
\affiliation{%
  \institution{Max Planck Institute for Intelligent Systems}
  \country{Germany}
}
\affiliation{%
  \institution{Mila, Université de Montréal}
  \country{Canada}
}

\author{Dimitrios Tzionas}
\email{d.tzionas@uva.nl}
\orcid{0000-0002-7963-2582}
\affiliation{%
  \institution{University of Amsterdam}
  \country{Netherlands}
}

\author{Michael J. Black}
\email{black@tuebingen.mpg.de}
\orcid{0000-0001-6077-4540}
\affiliation{%
  \institution{Max Planck Institute for Intelligent Systems}
  \country{Germany}
}

\renewcommand{\shortauthors}{Xiu, \etal}

\setcopyright{rightsretained}
\acmJournal{TOG}
\acmYear{2024} \acmVolume{43} \acmNumber{6} \acmArticle{} \acmMonth{12}\acmDOI{10.1145/3687771}

\begin{abstract}
    Generating \textbf{personalized} 3D avatars is crucial for %
    AR/VR.
    However, recent text-to-3D 
    methods %
    that generate 
    avatars for celebrities or fictional characters, %
     struggle with everyday people. 
    Methods for faithful reconstruction typically require full-body images in controlled settings. 
What if users could just upload their personal \UPDATE{``OOTD''} \UPDATE{(Outfit Of The Day)}photo collection and get a faithful avatar in return?
The challenge is that such 
\UPDATE{casual} 
photo collections contain diverse poses, challenging viewpoints, cropped views, and occlusion \UPDATE{(albeit with a consistent outfit, accessories and hairstyle)}. 
   We address this novel \mbox{\textbf{``Album2Human''}} task 
    by developing 
    \textbf{\modelname}, a novel model that generates %
    a faithful 3D avatar (in a canonical pose) 
    from a personal \UPDATE{OOTD} album, bypassing the challenging estimation of body and camera pose. 
    To this end, we fine-tune a foundational vision-language model (VLM) on 
    such photos,  
    encoding the appearance, identity, garments, hairstyles, and accessories of a person into separate learned tokens, instilling these cues into the VLM. 
    In effect, we exploit the learned tokens as 
    ``puzzle pieces" from which we assemble a faithful, personalized 3D avatar. 
    Importantly, we can customize avatars by simply inter-changing tokens. 
    As a benchmark for this new task, we create a new dataset, called \textbf{\dataname}, with 
    \UPDATE{41 subjects in a total of nearly 1k OOTD configurations}, 
    in  
     challenging partial photos with paired ground-truth 3D bodies.
    Evaluation shows that \modelname 
    not only has %
    high
    reconstruction accuracy, outperforming TeCH and MVDreamBooth, but also 
    a unique scalability to album photos, and demonstrating strong 
    robustness.  
    Our code and data are publicly available for research purpose at \page
\end{abstract}



\begin{teaserfigure}
    \centering
    \includegraphics[width=\textwidth]{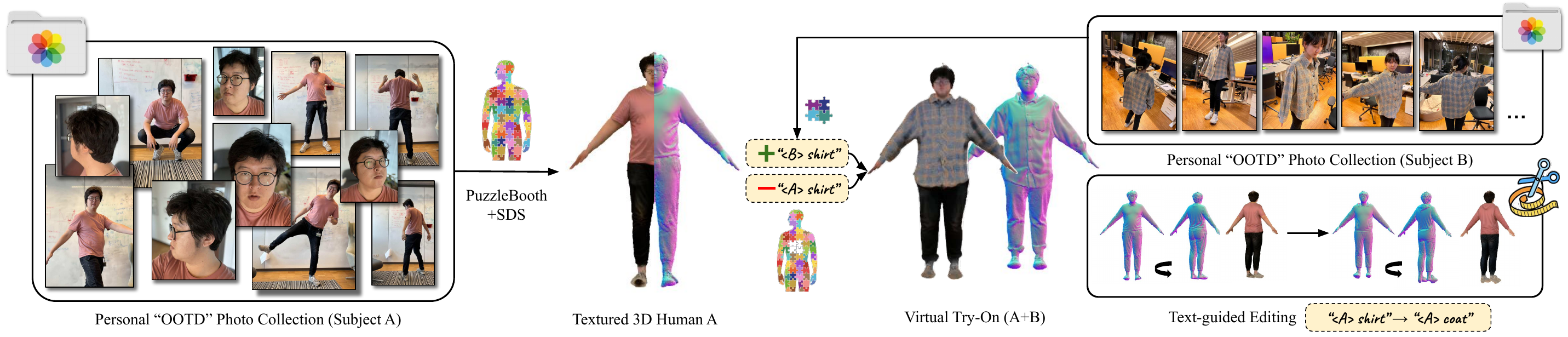}
    \caption{
        \modelname 
        reconstructs a 
        faithful, 
        personalized, 
        textured 3D human avatar from 
        \UPDATE{a personal}
        photo collection. 
        \UPDATE{That is, it takes as input a set of ``OOTD'' (Outfit Of The Day) personal photos with unconstrained body poses, camera poses, framing, lighting and backgrounds, albeit with a consistent outfit and hairstyle}. 
        All \UPDATE{these consistent factors} 
        are learned as separate unique tokens \mbox{\specific{<asset X>}} in a compositional manner, like pieces of a puzzle. 
        \modelname allows easily 
        inter-change tokens for downstream tasks, such as for 
        customizing avatars %
        and performing 
        virtual try-on \UPDATE{while preserving identity}, see \vid. %
    }
    \label{fig:teaser}
\end{teaserfigure}

\maketitle

\section{Introduction}

\newcommand{\mytextformat}{\itshape\epigraphsize}
\newenvironment{mytext}{\mytextformat}{}
\newenvironment{mysource}{\scshape\hfill}{}
\renewcommand{\textflush}{mytext} 
\renewcommand{\sourceflush}{mysource}
\let\originalepigraph\epigraph 
\renewcommand\epigraph[2]%
   {\setlength{\epigraphwidth}{\widthof{\mytextformat#1}}\originalepigraph{#1}{#2}}
\epigraph{In all chaos there is a cosmos, in all disorder a secret order.}{\textit{Carl Jung}}

\begin{figure*}%
    \includegraphics[width=\textwidth]{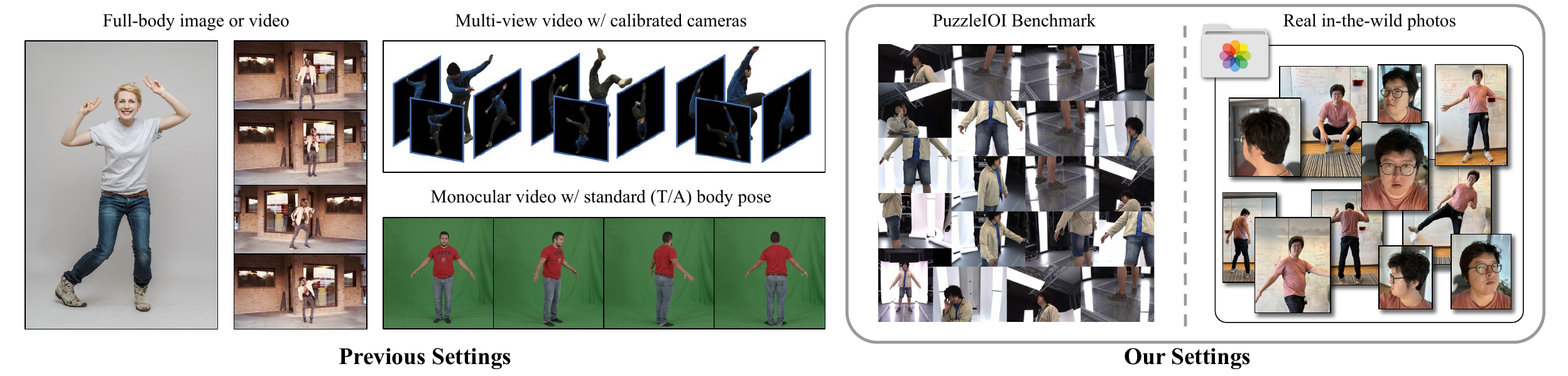}
    \caption{
        \UPDATE{\textbf{Image settings for avatar creation.}} 
        Past work (left) requires images with 
        full-body visibility, known camera calibration, or 
        simple 
        human poses. 
        \modelname operates on in-the-wild photos (right); \UPDATE{it assumes a consistent outfit, hairstyle and accessories, but deals with unconstrained human poses, camera settings, lighting and background.} 
        \UPDATE{Our \dataname dataset 
        contains 
        multi-view images with challenging crops paired with T-pose 3D ground truth.}
    }
    \label{fig:unconstrained}
\end{figure*}

Advances in text-guided digital human synthesis open the door to 3D avatar creation with arbitrary skin tones, clothing styles, 
\UPDATE{hairstyles and accessories}. 
While 
\CHECK{these advances have}
demonstrated great potential by generating iconic figures (such as Superman or Bruce Lee) and editing specific human features (such as wavy hair or full beards), the problem of crafting one’s \textit{personalized} full-body avatar is relatively unexplored.
Imagine that you are given a personal \UPDATE{``outfit of the day'' (OOTD)} photo album %
in casual snapshots: strolling through a park, crouching to tie a shoelace, seated at a cafe, \etc. These snapshots, capturing full-body actions, upper-body poses and close-up selfies with diverse backgrounds, \UPDATE{lighting  and camera settings}, form a rich photo 
collection. 
\UPDATE{Notably, this collection is relatively ``unconstrained'', that is, its only constraint is having a consistent identity, outfit, hairstyle and accessories, while every other factor can vary arbitrarily; \UPDATE{see \cref{fig:teaser}}}. Can we effectively construct from this %
album a personalized 3D avatar that vividly characterizes the user’s clothes, physique, and facial details? In this work, we investigate this novel task, which we call “\textbf{Album2Human}”, that transforms everyday album collections into textured 3D humans.

Compared to 
\UPDATE{work} 
that reconstructs general 3D %
\UPDATE{scenes} 
from %
photos %
with varying 
lighting conditions, cropping ratio, background and 
camera settings 
~\cite{martinbrualla2020nerfw,sun2022neuconw}, \mbox{\textbf{Album2Human}} is more challenging due to the additional factor of varying body articulation.
On the other hand, \textbf{Album2Human} drastically differs from prior work~\cite{alldieck2018videoavatar,peng2023implicit,10.1145/1661412.1618520} that creates personalized avatars from 
images 
captured in %
laboratory settings~\cite{tao2021function4d,Zheng2019DeepHuman,2023dnarendering,shen2023xavatar,xiong2023mvhumannet,isik2023humanrf,CAPE:CVPR:20}, in which full human bodies in limited body poses are captured using well calibrated and synchronized cameras with controlled lighting and simple backgrounds; \UPDATE{see \cref{fig:unconstrained}}.

While it is possible to create avatars from monocular (image or video) input 
as shown by some methods~\cite{deepcap,yang2023dif,xiu2022icon}, such methods perform poorly 
\UPDATE{for unusual body poses, motion blur, and occlusions,} 
because they rely on accurate 
\UPDATE{human and camera pose estimation} 
from full-body shots. 
Instead, \UPDATE{we bypass pose estimation}, and
\UPDATE{follow the new paradigm} 
of ``reconstruction as conditional generation'', 
as recently demonstrated for Text-to-Image (T2I) generation~\cite{huang2024tech,wu2023reconfusion,havefun,zhang2023humanref,gao2023contex}. 
Specifically, these works cast reconstruction from partial observations as ``inpainting'' unobserved regions through foundational-model priors, 
while imposing cross-view consistency. %
\UPDATE{We adapt existing T2I work~\cite{avrahami2023bas}} 
to learn \textbf{subject-specific} priors from a \UPDATE{\textbf{personal OOTD} image collection}, by finetuning T2I models on \UPDATE{such} images to 
capture \UPDATE{identity, pieces of clothing, accessories, and hairstyle into unique and inter-exchangeable tokens}, 
and extracting 3D geometry and texture with \sds based techniques~\cite{poole2022dreamfusion}.
\CHECK{Metaphorically, our model consumes ``unstructured'' data and digests this into a ``structured library''; that is, \textit{``seeking order in chaos, finding harmony in turmoil.''}}

Our insight to treat T2I models as personalized priors enables us to not only avoid \textit{explicit} per-pixel correspondences \CHECK{to a canonical human space}, but also to build avatars in a compositional manner.
To this end, 
given a photo collection of a person, 
various %
assets are extracted via an open-vocabulary segmentor~\cite{ren2024grounded}, such as the face, garments, accessories, and %
hairstyles. 
Each of these assets 
is labeled by %
a unique token as 
``\specific{<asset X>}''. 
We exploit these token-asset pairs, 
to finetune a pre-trained T2I model, 
so that it learns to generate 
\UPDATE{``personalized''} 
assets given a 
respective 
token.
\UPDATE{Based on this personalized T2I model, 
we produce} a 3D human avatar via Score Distillation Sampling (SDS) 
\UPDATE{given} 
a descriptive and compositional text prompt, \eg, ``\sloppy\specific{a DSLR photo of a man, with <asset1> face, wearing <asset0> shirt, \dots}'' 
(see~\cref{fig:teaser}). 
Here, each unique asset is like a puzzle piece, 
\UPDATE{characterizing the identity, hairstyle and dressing style} 
of the person. 
\UPDATE{In a sense, the learned tokens are used as puzzle pieces to assemble avatars, guided by text prompts.} 
Thus, we call our method \textbf{``\modelname''}.

Since there exists no benchmark 
for our 
new \textbf{Album2Human} task, %
we collect a new dataset, 
called \dataname, \UPDATE{of 41 subjects in a total of roughly 1k configurations (outfits, accessories, hairstyles).} 
Our evaluation metrics include both \emph{3D reconstruction errors} (\eg, Chamfer distances, P2S distances) between reconstructed shapes and ground-truth 3D scans, as well as \emph{2D image similarity measures} (\eg, PSNR, SSIM) between rendered multi-view images of the reconstructed surface and ground-truth textured scans.
Our \modelname is compatible with different types of diffusion models. 
We evaluate this on \dataname using two %
diffusion models, namely single-view Stable Diffusion ~\cite{rombach2022high} and multi-view \mvd~\cite{shi2023MVDream}. 
\UPDATE{Moreover}, we evaluate the contribution of each model component both qualitatively and quantitatively with an in-depth ablation analysis (\cref{sec:ablation}).

\vspace{2mm}
In summary, here we make the following main contributions:
\vspace{2mm}

\noindent
    \textbf{Task:} 
    We introduce a novel task, called ``\mbox{Album2Human}'', for reconstructing a 
    3D avatar 
    from 
    a \UPDATE{personal} \UPDATE{photo album with a consistent outfit, hairstyle and accessories, but unconstrained human pose, camera settings, framing, lighting and background}. 

\noindent
    \textbf{Benchmark:} 
    To benchmark this ``Album2Human'' task, 
    we collect a new dataset, called \dataname, \UPDATE{with challenging cropped images and paired 3D ground truth}. 
    This facilitates quantitatively evaluating methods 
    on both 3D reconstruction and view-synthesis quality. 

\noindent
    \textbf{Methodology:}
    \modelname 
    follows the 
    \UPDATE{fresh}
    paradigm of 
    \UPDATE{``reconstruction as conditional generation'', that is, it}
    performs \CHECK{implicit human canonicalization using a personalized T2I model to bypass}
    explicit 
    pose 
    estimation, or %
    re-projection pixel losses. 
    
\noindent
    \textbf{\CHECK{Analysis}:} 
    We conduct detailed evaluation and ablation studies to analyze the effectiveness and scalability of \modelname and each of its components, shedding light on potential future directions.
    
\noindent
    \textbf{Downstream applications:} 
    We show that \modelname's \UPDATE{highly-modular tokens}
    and 
    text guidance facilitates %
    \UPDATE{downstream}
    tasks \UPDATE{through two examples: } character editing and virtual try-on. 

Please check out more qualitative results and demos of applications in our \vid. \modelname is a step towards personalizing 3D avatars. 
To democratize this, code and \dataname dataset are public for research purpose at \page

\section{Related Work}

\zheading{3D Human Creation}
\UPDATE{Many works have explored how to reconstruct clothed humans from visual cues like multi-view images~\cite{saito2019pifu,lin2024fasthuman,peng2023implicit} or full-shot monocular video~\cite{alldieck2018videoavatar,alldieck2018dv,wengHumanNeRFFreeViewpointRendering2022,li2020monoport}. Recently, 
a lot of works strive to create human avatars characterized by language.} %
Initial work guided by language uses a CLIP embedding ~\cite{hong2022avatarclip} to sculpt coarse body shape. 
Recent work ~\cite{cao2023dreamavatar,huang2023dreamwaltz,liao2023tada,kolotouros2023dreamhuman,wang2023disentangled} captures finer geometry and texture for a clothed human, or multiple humans, by exploiting 
large-scale text-to-image models and 
Score Distillation Sampling (SDS)~\cite{wang2022score,poole2022dreamfusion}. 
In addition to text, %
when subject images are available, they are used %
to finetune the pretrained model~\cite{ruiz2022dreambooth} and to encourage fidelity via re-projection losses~\cite{huang2022one,huang2024tech,gao2023contex,havefun}. 
While %
SDS frameworks 
typically take a few thousand iterations, other work~\cite{chen2024ultraman} speeds up the process by one-step generation conditioned on a given image input.

However, all %
image-conditioned
methods 
assume reliable human pose estimation~\cite{pavlakos2019expressive} as a proxy representation to draw correspondences between the input image and the reconstructed 3D avatar. 
Hence, they require images with clean backgrounds, common body poses, and full-body views without crops. %
Furthermore, external controllers (\eg, ControlNet~\cite{zhang2023adding}, Zero123~\cite{liu2023zero1to3}) and additional geometric regularizers (\eg, Laplacian and Eikonal~\cite{chen2023fantasia3d}) appear essential to achieve high-quality output. In contrast, \modelname does not require any of these, thus, it is uniquely capable of operating on unconstrained personal-album photos. 
\camera{The most related work to ours is AvatarBooth~\cite{zeng2023avatarbooth}. It encodes personal identity (face, haircut) and overall outfit using two separate tokens \specific{[V]} and \specific{[W]}, while our method encodes each element (garments, haircut, face, accessories) with separate tokens. Additionally, we fine-tune one unified PuzzleBooth instead of separate DreamBooth for each part, making our method more efficient and scalable as the number of parts increases. More importantly, since our method treats the full-outfit as compositional tokens, it allows for 1) swapping parts of the clothing, see~\cref{fig:teaser}, and 2) training with heavily truncated photos using only visible tokens, see OOTD photos in~\cref{fig:unconstrained}.}

\zheading{Pose-Free Reconstruction in the wild}
In our work, the term ``pose'' refers not only to camera pose but also to body articulation. %
Camera pose plays a crucial role in 3D reconstruction, as it ``anchors'' 3D geometry onto 2D images~\cite{mildenhall2021nerf}, however, estimating it for in-the-wild images %
is highly challenging. 
Thus, to account for camera estimation errors, some work 
leverages joint optimization between the object and camera~\cite{wang2021nerf,xia2022sinerf,lin2021barf}, off-the-shelf geometric cue estimates~\cite{bian2023nope,fu2023colmap,meuleman2023progressively}, 
or learning-based camera estimation~\cite{wang2023pflrm,dust3r2023,zhang2024raydiffusion}. 
Body pose is also hard to estimate from in-the-wild images and is much higher dimensional than camera pose. 
Some work can reconstruct static scenes 
from in-the-wild images with challenging illumination conditions and backgrounds \cite{sun2022neuconw,martinbrualla2020nerfw}, but these cannot be applied to articulated objects, like humans. In our work, we tackle all above challenges for ``pose-free'' 3D human reconstruction.  That is, we tackle in-the-wild photos with unknown camera poses, unknown body poses, possibly truncated images (\eg headshots), and diverse backgrounds and illumination conditions, which are highly challenging for existing methods. Total-Selfie~\cite{chen2023total} takes daily selfies to generate a full-body 2D selfie, which aligns with our goal but not in the 3D domain.

\zheading{Large Vision-Language Models}
Large %
foundation models have achieved great progress in visual understanding~\cite{li2022blip,clip,kirillov2023segment} and  generation~\cite{rombach2022high,videoworldsimulators2024,SINC:2023}. 
As they are trained on a tremendous amount of data, their strong generalizability can be exploited for downstream tasks. 
In particular, subject-driven generation is fundamentally revolutionized by instilling the subject prior through fine-tuning the VLM models ~\cite{ruiz2022dreambooth,avrahami2023bas,tang2024realfill}, and Score-Distillation-Sampling techniques stand out~\cite{poole2022dreamfusion,wang2022score} for distilling ``common knowledge" from text-to-image models towards creating 3D objects. 
Work on model customization injects new concepts via fine-tuning (partial or whole) pre-trained networks~\cite{jain2021dreamfields,ruiz2022dreambooth,raj2023dreambooth3d,avrahami2023bas,kumari2022multi,boft}. 
Other work re-purposes the diffusion models %
to new tasks~\cite{ke2023repurposing,kocsis2024intrinsic,fu2024geowizard,ye2024stablenormal}.
\modelname leverage all the above techniques for faithful 
3D human-avatar 
generation 
from 
natural
images, a challenging task 
involving
widely varying appearance, 
lighting, 
backgrounds, body and camera poses.

\begin{figure*}%
    \includegraphics[width=\textwidth]{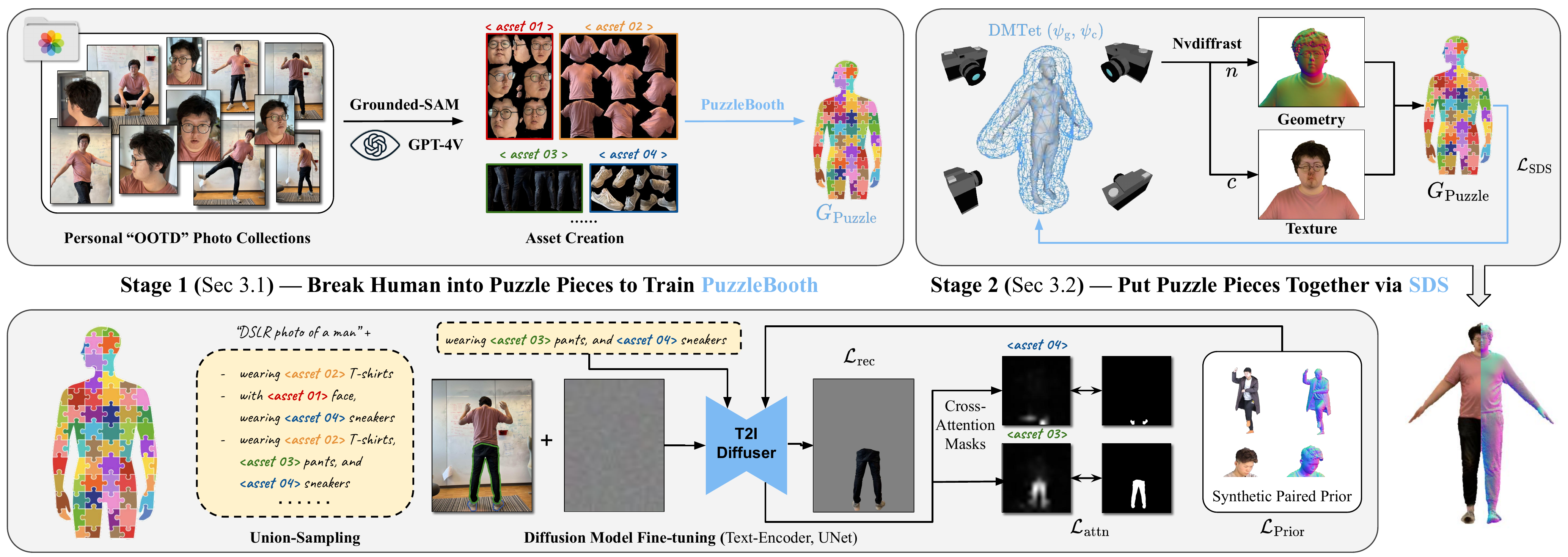}
    \caption{
        \textbf{\CHECK{Overview of \modelname.}}
        \UPDATE{The upper figure} shows the two main stages: 
        (1)
        \textit{\subsec} (\cref{sec:puzzlebooth}), where the unconstrained photo collections are captioned and segmented to create personalized puzzle pieces, 
        for training 
        PuzzleBooth ($G_\text{puzzle}$), and 
        (2)
        \textit{Create-3D-Avatar} (\cref{sec:sds}), where the T-posed textured tetrahedral body mesh
        is optimized using a multi-view SDS loss, $\mathcal{L}_\mathrm{SDS}$ (\cref{eq:nsfd}). 
        The bottom figure illustrates the training details of PuzzleBooth; 
        the Text-Encoder and the UNet of T2I Diffuser (\ie, Stable Diffusion) 
        \UPDATE{are} 
        fine-tuned using the 
        masked diffusion loss, $\mathcal{L}_\mathrm{rec}$ (\cref{eq:rec_loss}), 
        cross-attention loss, $\mathcal{L}_\mathrm{attn}$ (\cref{eq:attention_loss}), and 
        prior preservation loss, $\mathcal{L}_\mathrm{prior}$ (\cref{eq:prior_loss}). 
        Components marked in \textcolor{LightBlue}{light blue} are trainable or optimizable.}
    \label{fig:pipeline}
\end{figure*}

\section{Method}

Given an image collection $\{\mathcal{I}_1, \mathcal{I}_2, \dots \mathcal{I}_N\}$ of a person \camera{with the same outfit and style, or to say ``OOTD'' (Outfit Of The Day)},  we aim to build a %
3D avatar that captures the person's \CHECK{geometry} $\psi_g$ and appearance $\psi_c$. 
Notably, %
personal 
daily-life %
photos 
are unconstrained \UPDATE{(see \cref{fig:unconstrained})} as humans (1) appear in diverse poses and scales, (2) are often occluded or largely truncated, and (3) are captured from unknown viewpoints in diverse backgrounds. 
Thus, camera calibration and pose canonicalization for these photos are extremely challenging, making direct reconstruction of human avatars difficult.

Our key insight is to circumvent estimating human body poses and cameras, and, instead, to perform implicit human canonicalization via a foundation vision-language model (\eg, Stable Diffusion~\cite{rombach2022high}). 
\UPDATE{Our method is summarized visually in \cref{fig:pipeline}}, and has two main stages. 
Specifically, we first ``decompose'' %
photos into multiple %
assets (\eg, garments, accessories, faces, hair), all of which are linked with unique learned tokens by a personalized T2I model, PuzzleBooth (Sec.~\ref{sec:puzzlebooth}, $G_\text{puzzle}$ in~\cref{fig:pipeline}). Then, we ``compose'' these assets into a 3D full-body representation $\psi_g, \psi_c$ via \sds (Sec.~\ref{sec:sds}).

\subsection{PuzzleBooth -- Personalized Puzzle Pieces}
\label{sec:puzzlebooth}

Our first step is to segment subject images into multiple assets representing different human parts such as shirts and face. %
While one could build each asset individually, we adapt the %
``Break-A-Scene'' \cite{avrahami2023bas} approach, which shows that jointly learning multiple concepts 
significantly boosts performance, 
possibly because 
this 
facilitates global reasoning %
when multiple regions are simultaneously generated. %
Such a strategy is even more beneficial in our setting since human-related concepts, such as hairstyles, and dresscode, are harder to learn as their properties are correlated compared to clearly distinct objects in the setting of ``Break-A-Scene.''

\zheading{Asset Creation}
All %
images are segmented into $K$ assets, $\{{[\mathcal{V}_k]}\}_{k=1}^K$, each of which is associated with a segmentation mask, $\mathcal{M}_k$, a dedicated %
learnable token, \textcolor{blue}{[$v_k$]}, and its class label, \textcolor{red}{$[c_k]$}, such as ``pants'' or ``skirt.'' 
In addition, we also obtain a coarse view direction, \textcolor{GreenColor}{$d$}, for each image. 
All such information is obtained automatically by \mbox{Grounded-SAM}~\cite{ren2024grounded} and \mbox{GPT-4V} (see prompt at~\cref{sec:prompt-sup}). 
Specifically, we query GPT-4V with an image to directly get the property of each asset ,\textcolor{red}{$[c_k]$}, and coarse view direction, \textcolor{GreenColor}{$d$}. 
Then, given the full list of queried asset names $\{{\color{red}{[c_k]}}\}_{k=1}^K$, Grounded-SAM outputs segmentation masks if they are present. 

\zheading{Two-Stage Personalization} 
We finetune a pretrained text-to-image diffusion model~\cite{rombach2022high,shi2023MVDream} so that it adapts to the new assets. 
Following ``Break-A-Scene'' \cite{avrahami2023bas},
we firstly optimize 
the ``text'' part for 1,000 iterations, \ie, the CLIP embedding of \CHECK{asset} 
token \textcolor{blue}{[$v_k$]}, and then 
the ``visual'' part for 4,000 iterations, \ie, the weights of the UNet, 
in two stages:
In the first stage, only CLIP embedding of the \CHECK{asset} tokens \textcolor{blue}{[$v_k$]} are optimized with a large learning rate. 
In the second stage, both the ``text'' and ``visual'' part are optimized with a small learning rate. 
This strategy effectively prevents guidance collapse~\cite{gao2024graphdreamer} between newly introduced tokens \textcolor{blue}{[$v_k$]} and existing asset names \textcolor{red}{[$c_k$]}, 
or, equivalently, preserves the compositionality of visual concepts.

During training, we randomly select, for every image $\mathcal{I}$, a subset of $J<K$ assets that appear in the image and train the model on the union set of these selected assets. This union sampling strategy, originally introduced in ~\cite{avrahami2023bas}, is crucial for effective asset disentanglement.
Specifically, the \emph{mask union} is done via a pixel-wise union operation, $\mathcal{M}_\cup = \cup_{i=1}^{J}\mathcal{M}_{i}$, while the \emph{image union} 
\UPDATE{applies the union mask on the image}, 
$\mathcal{I}_\cup = \mathcal{I} \odot \mathcal{M}_\cup$. The union text prompt, $p_\cup$, is constructed by concatenating selected assets, where \textcolor{blue}{[$v_{1 \sim m}$]} and \textcolor{red}{[$c_{1 \sim m}$]} denote the facial features, and \textcolor{blue}{[$v_{m+1 \sim j}$]} and \textcolor{red}{[$c_{m+1 \sim j}$]} denote the features of garments or accessories. 
Here shows a prompt template of $p_\cup$, ``\specific{a high-resolution DSLR color image of a man/woman with \textcolor{blue}{[$v_{1}$]} \textcolor{red}{[$c_{1}$]}, \dots, \textcolor{blue}{[$v_{m}$]} \textcolor{red}{[$c_{m}$]}, and wearing \textcolor{blue}{[$v_{m+1}$]} \textcolor{red}{[$c_{m+1}$]}, \dots, \textcolor{blue}{[$v_{j}$]} \textcolor{red}{[$c_{j}$]}, \textcolor{GreenColor}{[$d$]} view}''. 

\zheading{Losses} In both optimization stages, the model is trained to encourage concept separation while still retaining its generalization capability.  
To do so, the model is optimized with three loss terms: a 
Masked Diffusion   Loss, $ \mathcal{L}_\text{rec}$, 
Cross-Attention    Loss, $\mathcal{L}_\text{attn}$, and 
Prior Preservation Loss, $\mathcal{L}_\text{prior}$. 
The overall training objective is 
$    \mathcal{L}_\text{total} = \mathcal{L}_\text{rec} + \lambda_\text{attn}\mathcal{L}_\text{attn} + \mathcal{L}_\text{prior}
$    
where $\lambda_\text{attn}=0.01$. 

The \textit{Masked Diffusion Loss} encourages fidelity in replicating each concept 
\UPDATE{via a}
pixel-wise reconstruction within the segmented mask:
\begin{equation}
    \mathcal{L}_\text{rec} = \mathbb{E}_{z, \epsilon \sim \mathcal{N}(0, 1), t }\big[ \Vert [ \epsilon - \epsilon_\theta(z_{t},t,p_\cup) ] \odot \mathcal{M}_\cup \Vert_{2}^{2}\big],
    \label{eq:rec_loss}
\end{equation}
where $\mathcal{M}_\cup$ is the union mask, and $\epsilon_\theta(z_{t},t,p_\cup)$ is the denoised output at timestep $t$ given the union prompt, $p_\cup$, and the visual feature, $z_{t}$.

For disentanglement purpose, we use a 
\textit{Cross-Attention Loss} ~\cite{avrahami2023bas} to encourage each of the newly-added tokens \UPDATE{to be} exclusively associated with only the target asset: 
\begin{equation}
    \mathcal{L}_{\mathrm{attn}} = \mathbb{E}_{z, j, t}\big[ \Vert \mathcal{CA}_{\theta}(\textcolor{blue}{v_j}, z_t) - \mathcal{M}_{j} \Vert_{2}^{2}\big],
    \label{eq:attention_loss}
\end{equation}
where $\mathcal{CA}_{\theta}(\textcolor{blue}{v_j}, z_t)$ is the cross-attention map in the diffusion U-Net between the newly-added token, \textcolor{blue}{[$v_j$]}, and the visual feature, $z_{t}$. 

\begin{figure}[!htbp]
    \includegraphics[width=\linewidth]{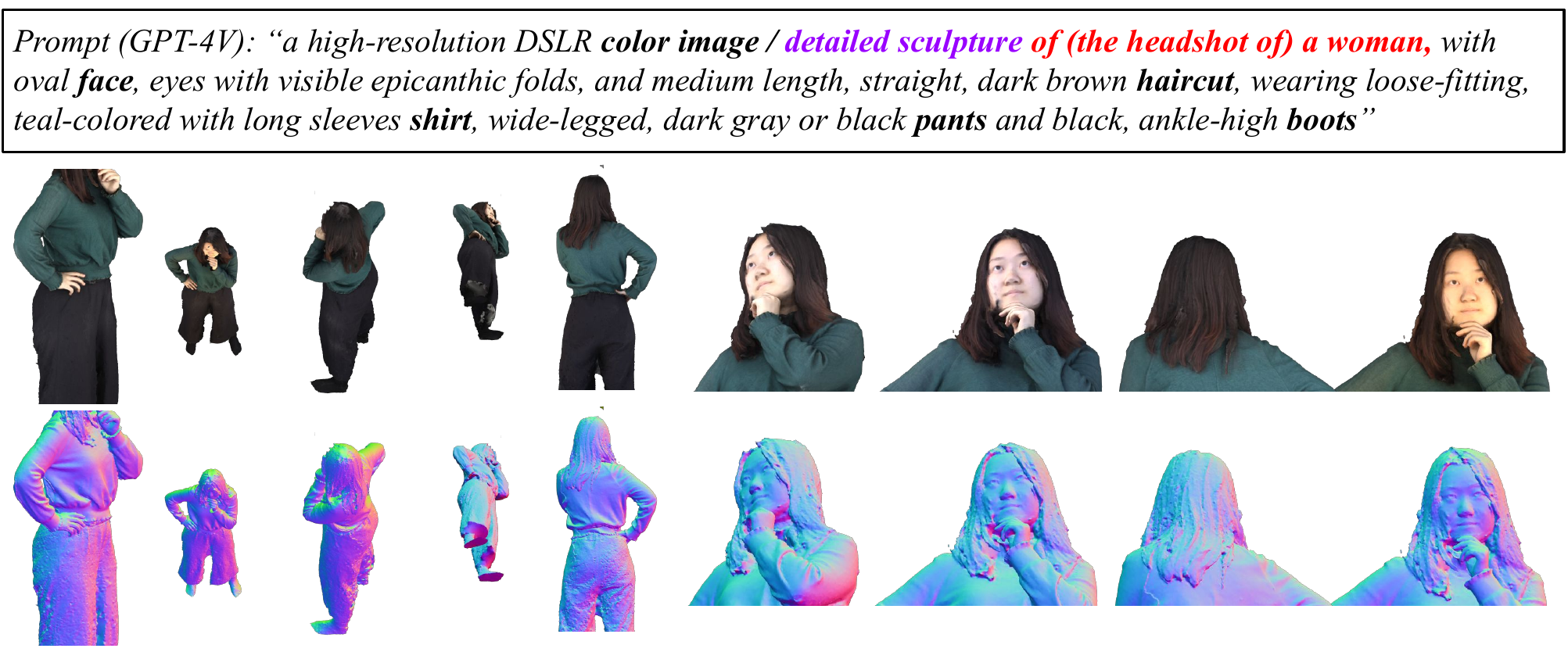}
    \caption{
        \textbf{\CHECK{Color-Normal Synthetic Prior.}}
        The descriptions are generated via GPT-4V~\cite{gpt4v}, where the prompt of the RGB image starts with ``\specific{a high-resolution DSLR colored image}'', while that of the normal image starts with ``\specific{\textcolor{PurpleColor}{a detailed sculpture of}}'' 
        The zoomed-in head images are generated by appending ``\specific{\textcolor{red}{the headshot of}}''.}
    \label{fig:synthetic-data}
\end{figure}

Lastly, we apply a \textit{Prior Preservation Loss}~\cite{ruiz2022dreambooth} to retain the generalization capability of the vanilla T2I model --- Stable Diffusion (SD-2.1). 
The model is trained to reconstruct images with general concepts when the special tokens are removed from prompts.
General human images come from two sources: 
(1) 
\CHECK{Generated images}, $\mathcal{I}^\mathrm{pr}_\mathrm{gen}$, \CHECK{from SD}. 
(2)
Synthetic color-normal pairs (see~\cref{fig:synthetic-data}), $\mathcal{I}^\mathrm{pr}_\text{syn}$,  rendered from multiple views, \UPDATE{from \mbox{THuman2.0}}~\cite{tao2021function4d}. 
The latter is to improve the geometry quality and color-normal consistency~\cite{huang2023humannorm}. 
Instead of applying the prior preservation loss for individual concepts separately, we find it beneficial to compute the loss on the entire human images. 
\begin{equation}
\mathcal{L}_\text{prior} = \mathbb{E}_{z^\mathrm{pr}, \epsilon \sim \mathcal{N}(0, 1), t }\big[ \Vert [ \epsilon - \epsilon_\theta(z^\mathrm{pr}_{t},t,p^{*}_\cup) ] \Vert_{2}^{2}\big] \label{eq:prior_loss}    
\end{equation}
where $p^{*}_\cup$ is the text prompt without special tokens.

\subsection{\modelname ~-- %
Putting the Puzzle Pieces Together}
\label{sec:sds}

With the fine-tuned diffusion model customized for all provided assets, we are able to distill a descriptive 3D avatar via SDS. 

\medskip

\zheading{%
Score Distillation Sampling (SDS)} 
A diffusion-based generative model, $D(\mathrm{z})$, where $\mathrm{z} \sim \mathcal{N}(0,\mathrm{I})$, captures the real image distribution. SDS~\cite{poole2022dreamfusion} is a technique that guides the parameterization of images, $\mathrm z(\psi)$, where $\psi$ denotes neural networks, to generate images that better align with control signals, such as text. 
The core idea is to approximate the parameter gradient $\nabla_{\psi}\mathcal{L}$ as a weighted reconstruction residual. As the vanilla method suffers from color oversaturation, we use an improved SDS 
-- Noise-Free Distillation Sampling (NFDS)~\cite{katzir2024noisefree}. 
This modifies the guidance from \UPDATE{a single} 
reconstruction residual into two composed residual terms $\delta_C \text{ and } \delta_D$. 
Specifically, by denoting the derived gradient of a network $\psi$ from NFSD  as $ \nabla \mathcal{L}_{\text{NFDS}}(\mathrm z, \psi)$:

\begin{equation}
    \nabla_{\psi}\mathcal{L}_{\text{NFDS}}(\mathrm z, \psi) = w(t) (\delta_D + s\delta_C) \frac{\partial \mathrm z} {\partial \psi} \text{, \quad \UPDATE{where}}\\
    \label{eq:nsfd}
\end{equation}

\begin{equation}
\begin{aligned}
 \delta_\mathrm{C} (z_t, p, t) &= 
        \epsilon_\theta (z_t; p, t) - \epsilon_\theta (z_t; \varnothing, t),\\
    \delta_\mathrm{D} (z_t, t) &= 
    \begin{cases}
        \epsilon_\theta (z_t; \varnothing, t), & \text{if } t \le 200 \\
        \epsilon_\theta (z_t; \varnothing, t) - \epsilon_\theta (z_t; p^\mathrm{neg}, t), & \text{otherwise},
    \end{cases}
\end{aligned}
\end{equation}

 In our case, $\mathrm z$ is the (latent of) diffusion output (human images or normals) and $\psi$ \UPDATE{denotes the} 
 3D avatar representation \UPDATE{(both $\psi_g$,  $\psi_c$)}, $s$ is the guidance scale. We follow NFDS and set $s = 7.5$.

\zheading{Representation and Initialziation}
The 3D human is parameterized with DMTet~\cite{shen2021deep, gao2020learning}, a flexible tetrahedron-based 3D neural representation. 
\UPDATE{The geometry, $\psi_\mathrm{g}$, and appearance, $\psi_\mathrm{c}$, are optimizable}, and 
can be differentially rendered into normal, $\mathrm{n}$, and colored images, $\mathrm{c}$. 
The geometry $\psi_g$ is first initialized to an A-posed SMPL-X body~\cite{pavlakos2019expressive}. 

\zheading{Optimization}
We use the full-text description of the human $p^\text{all}$ as 
\UPDATE{a guiding} 
prompt. It is a concatenation of text prompts from all assets  \ie, $ (\textcolor{blue}{v_i},\textcolor{red}{c_i}), \dots, (\textcolor{blue}{v_K},\textcolor{red}{c_K})$. 
We optimize geometry and color separately in two optimization stages for 10,000 iterations each, both using Noise-Free-Score Distillation (NFSD). 
In the first stage, the avatar's geometry is guided in the surface normal space, $\nabla \mathcal{L}^{\text{norm}} \equiv \nabla \mathcal{L}_{\text{NFDS}}[\mathrm n, \psi_g]$. We additionally prepend ``\specific{a detailed sculpture of}'' to the full-text to indicate the guidance space.  In the second stage, its appearance is guided by $\nabla \mathcal{L}^{\text{color}} \equiv \nabla \mathcal{L}_{\text{NFDS}}[\mathrm c, \psi_c]$. The camera settings for multi-view SDS are in ~\cref{sec:camera-sup}, \camera{and see the detailed negative prompt $p^\text{neg}$ in~\cref{sec:negative-prompt}.}

\section{Experiments}

It has been a long-standing challenge in the field of ``Text-to-3D'' (including ``Text-to-Avatar'') to \textit{quantitatively} benchmark new algorithms. 
Existing benchmarks %
are typically less reliable than ours, because they sample 3D avatars from a relatively small collection of prompts and evaluate the quality of these avatars 
\UPDATE{through} 
perceptual studies with 
\UPDATE{a limited number of participants}.

While \modelname adopts the 
``Text-to-3D'' paradigm, 
its goal %
is to reconstruct avatars from photos of 
\UPDATE{a specific person in a specific outfit}, 
rather than to randomly
generate 
avatars. 
As a result, a natural and reliable way to benchmark \modelname is to 
\UPDATE{exploit a 4D scanner (with synced IOI color cameras\footnote{\url{https://www.ioindustries.com/cameras}}) 
for capturing ground-truth 3D shape and appearance}, 
and to measure the reconstruction \UPDATE{error} %
\UPDATE{between the reconstructed and ground-truth shape and appearance}.
We thus build a dataset, called \dataname (\cref{sec:dataset}), on which we 
\UPDATE{evaluate \modelname and ablate its %
components.} 

\subsection{\dataname Dataset}
\label{sec:dataset}

We create \dataname \UPDATE{(see statistics in \cref{tab:puzzleioi})} to simulate real-world album photos of humans, which: 
(1) 
cover a wide range of human identities (\CHECK{\textbf{\#ID}} \UPDATE{column in \cref{tab:puzzleioi}}) 
and 
daily outfits 
(\textbf{\#Outfits}, \camera{representing the total number of outfits with the specific number varying among 15\textasciitilde46 for different ID}),
(2) 
span numerous views (\textbf{\#Views}) to mimic real-world captures (\eg,  occlusion, out-of-frame cropping), 
and
(3)
include text descriptions (\textbf{Text}), %
and 
ground-truth textured A-posed scans (\textbf{Scan, Texture}) and their SMPL-X fits (\textbf{SMPL-X}) for shape initialization purposes.

\begin{table}[t]
\scriptsize
\centering
\renewcommand{\arraystretch}{1.26}
\setlength{\tabcolsep}{0.5pt}
\caption{\textbf{Datasets related to \dataname.} ``--'' means \CHECK{image} captures are unavailable. ``Scan'' is A-posed, and ``SMPL-X'' is its respective SMPL-X fit.}
\begin{tabular}{llcccc|cccc}
\textbf{Dataset} & Reference & \textbf{\#Views} & \textbf{\#ID} & \textbf{\#Outfits}  & \textbf{\#Actions } & \textbf{SMPL-X} & \textbf{Scan} & \textbf{Text} & \textbf{Texture} \\\shline
ActorsHQ & \cite{isik2023humanrf}  & 160 & 8 & 8 & 52 & \cmark & \cmark & \xmark & \cmark\\
MVHumanNet & \cite{xiong2023mvhumannet} & 48 & 4500 & 9000 & 500 & \cmark  & \xmark & \cmark & \cmark\\
HuMMan & \cite{cai2022humman} & 10 & 1000 & 1000  & 500 & \xmark  & \xmark & \xmark & \cmark\\
DNA-Rendering & \cite{2023dnarendering} & 60 & 500 & 1500 & 1187 & \xmark  & \xmark & \cmark & \xmark\\
THuman2.0 & \cite{tao2021function4d} & -- & 200 & 500 & -- & \xmark  & \xmark & \xmark & \cmark\\
CAPE & \cite{CAPE:CVPR:20}  & -- & 15 & 8 & 600  & \cmark & \cmark & \xmark & \xmark\\
BUFF & \cite{shape_under_cloth:CVPR17}  & -- & 5 & 2 & 3 & \cmark  & \cmark & \xmark & \cmark\\
\hline
\dataname (Ours) & {} & 22 & 41 & 933 & 40 & \cmark & \cmark & \cmark & \cmark\\
\end{tabular}
\label{tab:puzzleioi}
\end{table}

\zheading{A-Pose SMPL-X \& Scan} 
Almost all ``Text-to-Avatar'' methods~\cite{liao2023tada,huang2023humannorm,kolotouros2023dreamhuman,cao2023dreamavatar,yuan2023gavatar} use an \mbox{A-pose} body for shape initialization due to its minimal self-occlusions. %
Thus, 
we adhere to this empirical setting in \dataname. 
For each subject \textbf{(ID+Outfit)}, we capture a \UPDATE{ground-truth} \mbox{A-posed} 3D scan, and register a \mbox{SMPL-X} model to it as in AGORA~\cite{patel2020agora}. 

\zheading{Multiple Views}
To simulate the diversity and imperfections of real-world photos, for each subject \textbf{(ID+outfit)} we randomly sample 120 photos from each multi-view human action sequence (approx. 760 frames / subject) that we capture with 22 
cameras; \UPDATE{see} \cref{fig:unconstrained}. 
The captured %
images are %
segmented and shuffled to build the \CHECK{training dataset for \mbox{PuzzleBooth}} (\cref{sec:puzzlebooth}).

\zheading{Text Description}
Similar to how image caption is done in~\cref{sec:puzzlebooth}-\textbf{Asset Creation}, here we randomly select two frontal full-body images and use GPT-4V to query the \CHECK{asset names} and corresponding descriptions of visible assets. 
We use the position of the groundtruth camera to categorize the photos into 4 view groups \specific{\{front, back, side, overhead\}} in \dataname, while we use GPT-4V to automatically label viewpoints from in-the-wild images.

\subsection{2D and 3D Metrics}
\label{sec:metrics}

We conduct quantitative evaluation on the \dataname dataset (Sec.~\ref{sec:dataset}). 
To 
evaluate the quality of \textbf{shape reconstruction} 
we report 
three metrics: 
\textbf{Chamfer distance} (bidirectional point-to-surface, \textit{cm} as unit), 
\textbf{P2S distance} (1-directional point-to-surface, \textit{cm} as unit) distance, and 
\textbf{L2 \UPDATE{error for} \textbf{Normal} maps} 
\UPDATE{rendered for four views} 
($\{0^{\circ}, 90^{\circ}, 180^{\circ}, 270^{\circ}\}$) 
\CHECK{to capture} 
local surface details. 

To evaluate the 
\UPDATE{quality of} 
\textbf{\UPDATE{appearance reconstruction}}, 
we render multi-view color images as 
above, and report three image-quality metrics:
\textbf{PSNR}  (Peak Signal-to-Noise Ratio), 
\textbf{SSIM}  (Structural Similarity) and 
\textbf{LPIPS} (Learned Perceptual Image Path Similarity).

\subsection{Benchmark}
\label{sec:benchmark}
\modelname is a general 
\CHECK{framework}, %
compatible with different %
diffusion models. 
In \cref{table:benchmark} we benchmark variants of \modelname with %
two different backbones: 
(1) 
vanilla Stable Diffusion~\cite{rombach2022high}, \ie, \mbox{SD-2.1}~\footnote{\url{huggingface.co/stabilityai/stable-diffusion-2-1-base}}, 
and 
(2)
\mbox{MVDream}~\cite{shi2023MVDream}~\footnote{\url{huggingface.co/ashawkey/mvdream-sd2.1-diffusers}} fine-tuned from vanilla SD using multi-view images rendered from 
Objaverse~\cite{deitke2023objaverse}. 
The shared basic pipeline for \modelname, and the \sota image-to-3D methods 
TeCH~\cite{huang2024tech} and MVDreamBooth~\cite{shi2023MVDream} is: (1) finetune \CHECK{these backbones} with subject images and (2) later to optimize avatars with text-guided SDS optimization.

\zheading{Quantitative Evaluation}
\cref{table:benchmark} \UPDATE{shows that} \modelname is on par with TeCH on 3D metrics, while outperforming it on all 2D metrics. 
Note that, to enhance shape quality, TeCH employs multiple %
supervision signals and regluarization terms, 
including 
normal maps predicted from the input image via ECON~\cite{xiu2022econ}, 
silhouette masks produced by SegFormer~\cite{xie2021segformer} and a 
Laplacian regularizer. 
In terms of texture quality, TeCH uses an RGB-based chamfer loss to minimize color shift between the input image and the backside texture, while its front-side texture is achieved 
\CHECK{by back-projecting} 
the input image. 
In contrast, \modelname achieves on-par 3D accuracy and better texture quality \emph{\textbf{without}} %
any of these auxiliary losses, regularizers, or \CHECK{pixel back-projection}. %

As for the MVDream-based comparison, \modelname outperforms \mbox{MVDreamBooth} on texture quality by a large margin (PSNR \bgood{+10.09}, LPIPS \bgood{-8.79}), and on geometry quality 
(measured by Chamfer and P2S), while showing comparative performance with the baselines on normal consistency. \camera{All the methods in~\cref{table:benchmark} are implemented with the same distillation method (NFDS) and the same 3D representation (\dmtet). 
What distinguishes \modelname from \mbox{MVDreamBooth} is PuzzleBooth, our puzzle-wise training strategy.
} 
Without this, %
2D diffusion models fine-tuned on human photos with complex poses and cropping might 
\UPDATE{produce} 
completely flawed 3D humans, with low-quality (even full black) textures or overly smooth shapes; see~\cref{fig:qualitative}.

\begin{figure*}[!htbp]
\includegraphics[width=\linewidth]{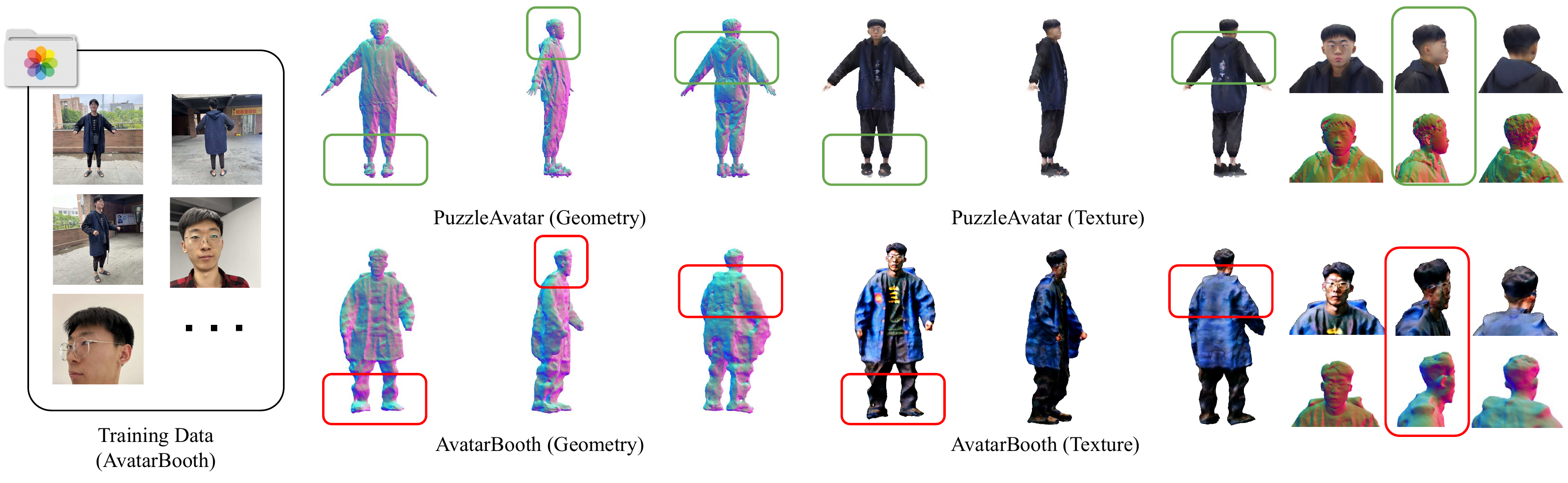}
\vspace{-2.0 em}
\caption{\textbf{AvatarBooth~\cite{zeng2023avatarbooth} vs. \modelname.} AvatarBooth overlooks the compositionality of garments and utilizes two separate DreamBooths (Head, Body) along with ControlNet, making it more complex and less scalable than \modelname.}
  \label{fig:avatarbooth}
\end{figure*}

\zheading{Qualitative Evaluation}
As depicted in~\cref{fig:qualitative}, \modelname has various advantages over TeCH: 
(1)
\emph{Enhanced front-back consistency}, because \modelname treat all views with ID-consistent generation, while TeCH introduces inconsistency between the front view created by reconstruction and the back view created by hallucination.
(2)
\emph{Reduced non-human artifacts}, \modelname bypass the dependence on numerous off-the-shelf estimators used in TeCH, \UPDATE{for which} 
non-human artifacts arise when segmentation or normal map estimation fails. 
(3)
\emph{Improved geometry-texture disentanglement}, where \modelname excels in separating 
shirt stripes 
compared to TeCH
This is mainly attributed to the failed normal-map estimation from the input image (see~\cref{fig:qualitative}, 3th row, rightmost normal estimate), which relies on often incorrectly estimated normal maps from the input image.
Notably, the comparison with \mbox{MVDreamBooth} highlights \modelname's ability in producing intricate geometric details and textures. 
We also compare with AvatarBooth~\cite{zeng2023avatarbooth} on the same photo collections, see results at~\cref{fig:avatarbooth} and \vid.

\begin{table*}[t]
\centering
\caption{\textbf{%
Evaluation on full \dataname (933 OOTD).}
\textdagger~means using \mbox{SMPL-X} fits of ground-truth scans to initialize DMTet and \UPDATE{factor} %
 out pose error (unlike the vanilla TeCH~\cite{huang2024tech} that estimates \mbox{SMPL-X} using \mbox{PIXIE}~\cite{feng2021pixie}).
 The best results are marked with ``\textbf{bold}''. 
 ``\textcolor{RedColor}{Ratio\%}'' is the relative performance drop, while ``\textcolor{GreenColor}{ratio\%}'' is the relative performance gain, \wrt the competitors, \ie TeCH and \mbox{MVDreamBooth}~\cite{shi2023MVDream}.
}
\scriptsize
\resizebox{0.9\linewidth}{!}{
\begin{tabular}{cc|ccc|ccc}
Method & Backbone &
  \multicolumn{3}{c|}{3D Metrics (Shape)} &
  \multicolumn{3}{c}{2D Metrics (Color)} \\
\multicolumn{2}{c|}{} &
  Chamfer $\downarrow$ &
  P2S $\downarrow$ &
  Normal $\downarrow$ & 
  PSNR$\uparrow$ &
  SSIM$\uparrow$ &
  LPIPS$\downarrow$\\ \shline
  TeCH\textsuperscript{\textdagger} & SD-2.1-base
  & 1.646 & \textbf{1.590} & \textbf{0.076} &  23.635 & 0.919 & 0.065
\\ 
\modelname & SD-2.1-base
   & \textbf{1.617} \bgood{-1.76} & 1.613 \bbad{+1.45} & 0.077 \bbad{+1.32} &  \textbf{24.687} \bgood{+4.45} & \textbf{0.930} \bgood{+1.20} & \textbf{0.062} \bgood{-4.62}
\\ 
\hline
MVDreamBooth\textsuperscript{\textdagger} &  MVDream 
   & 1.705 & 1.835 & 0.100 & 19.401 & 0.909 & 0.091
\\
\modelname &MVDream
   & 1.697 \bgood{-0.47} & 1.811 \bgood{-1.31} & 0.101 \bbad{+1.00} & 21.361 \bgood{+10.09} & 0.906 \bbad{-0.33} & 0.083 \bgood{-8.79}
\\  
  
\end{tabular}}
\label{table:benchmark}
\end{table*}

\begin{table*}[t]
\centering
\caption{\textbf{Ablation study on subset of \dataname (120 OOTD).}
 The best results are marked with ``\textbf{bold}'', the second best results are marked with and \underline{underline}. The ``\textcolor{RedColor}{ratio\%}'' is the relative performance drop, and ``\textcolor{GreenColor}{ratio\%}'' is the relative performance gain, \wrt \colorbox{Gray}{\modelname}, where the drop larger than 20\% are marked with ``\textbf{\textcolor{red}{bold}}''. Group-A summarizes the \textit{failed attempts}, Group-B justifies the \textit{key components}, and \camera{Group-C} analyses the \textit{scalability} of our method.
 } 
\scriptsize
\renewcommand{\arraystretch}{1.26}
\resizebox{0.9\linewidth}{!}{
\begin{tabular}{ll|lll|lll}
Group & Method & 
  \multicolumn{3}{c|}{3D Metrics (Shape)} &
  \multicolumn{3}{c}{2D Metrics (Color)} \\
\multicolumn{2}{c|}{} &
  Chamfer $\downarrow$ &
  P2S $\downarrow$ & Normal $\downarrow$ &
  PSNR$\uparrow$ &
  SSIM$\uparrow$ &
  LPIPS$\downarrow$\\ \shline
  & TeCH\textsuperscript{\textdagger} & 1.600 & \underline{1.541} & 0.073 & 23.665 & 0.919 & 0.065\\
\rowcolor{Gray} & \textbf{\modelname} &  \underline{1.589} & 1.570 & 0.075  & \underline{24.718} & \textbf{0.931} & \textbf{0.061}\\
\hline
 A. & w/ detailed GPT-4V description & 1.604 \bbad{+0.9} & 1.607 \bbad{+2.4} & 0.079 \bbad{+5.3} &  24.208 \bbad{-2.1} & \underline{0.929} \bbad{-0.2} & \underline{0.062} \bbad{+1.6}\\
\hline
  & w/o view prompt & 1.641 \bbad{+3.3} & 1.653 \bbad{+5.3} & 0.082 \bbad{+9.3} &  23.929 \bbad{-3.2} & 0.928 \bbad{-0.3} & 0.064 \bbad{+4.9}\\
 & w/o NFSD (vanilla SDS) & 1.624 \bbad{+2.2} & 1.604 \bbad{+2.2} & \underline{0.072} \bgood{-4.0} & 20.441 \bbad{-17.3} & 0.923 \bbad{-0.9} & 0.071 \bbad{+16.4}\\
 B. & w/o synthetic normal+color & 2.194 \textbf{\bbad{+38.1}} & 2.493 \textbf{\bbad{+58.8}} & 0.130 \textbf{\bbad{+73.3}} &  21.940 \bbad{-11.2} & 0.912 \bbad{-2.0} & 0.078 \textbf{\bbad{+27.9}}\\ 
 & w/o synthetic normal & 2.089 \textbf{\bbad{+31.5}} & 2.335 \textbf{\bbad{+48.7}} & 0.123 \textbf{\bbad{+64.0}}  & 23.684 \bbad{-4.2} & 0.919 \bbad{-1.3} & 0.074 \textbf{\bbad{+21.3}}\\
 & w/o synthetic color & 1.680 \bbad{+5.7} & 1.687 \bbad{+7.5} & 0.084 \bbad{+12.0}  & 23.813 \bbad{-3.7} & 0.927 \bbad{-0.4} & 0.063 \bbad{+3.3}\\
 \hline
 C. & multi-subject training (5 subjects / model) & 1.809 \bbad{+13.8} & 1.560 \bgood{-0.6} & 0.080 \bbad{+6.7} & \textbf{24.990} \bgood{+1.1} & \underline{0.929} \bbad{-0.2} & \underline{0.062} \bbad{+1.6} \\

 & w/o full-body images & 1.603 \bbad{+0.9} & 1.580 \bbad{+0.6} & 0.073 \bgood{-2.7} & 23.703 \bbad{-4.1} & \textbf{0.931} \btie{0.0} & \underline{0.062} \bbad{+1.6}\\
 & 50\% training data & 1.590 \bbad{+0.1} & 1.569 \bgood{-0.1} & 0.074 \bgood{-1.3} & 24.095 \bbad{-2.5} & 0.930 \bbad{-0.1} & \textbf{0.061} \btie{0.0}\\
  & 10\% training data & \textbf{1.583} \bgood{-0.4} & \textbf{1.531} \bgood{-2.5} & \textbf{0.069} \bgood{-8.0} & 23.477 \bbad{-5.0} & 0.928 \bbad{-0.3} & 0.062 \bbad{+1.6}\\

\end{tabular}}
\label{table:ablation}
\end{table*}

\subsection{Ablations}
\label{sec:ablation}

\begin{figure*}[!htbp]
  \includegraphics[width=\linewidth]{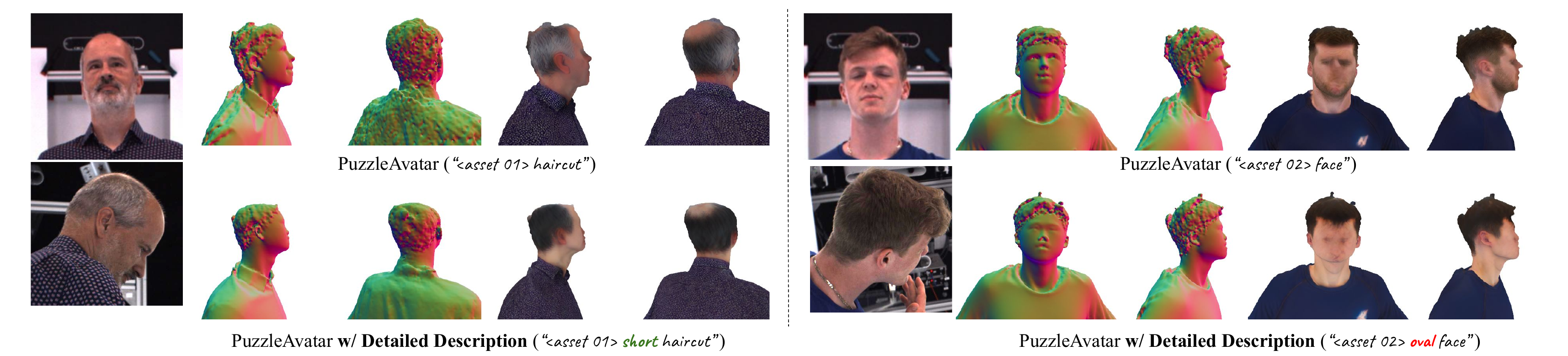}
  \vspace{-2.0 em}
  \caption{\textbf{\CHECK{Detailed vs. Plain Prompt. }} Token \textcolor{blue}{<asset X>} suffices to maintain the appearance of assets. Elaborate prompts could introduce bias and hallucination.}
  \label{fig:detailed}
\end{figure*}

\zheading{Ablation: Common Practices} 
In ~\cref{table:ablation}-B, we analyze the effect of common practices that have been shown to be beneficial for general scenes, including view-specific prompts~\cite{ruiz2022dreambooth}, NFSD over vanilla SDS~\cite{katzir2024noisefree}, and a prior preservation loss~\cite{ruiz2022dreambooth,huang2023humannorm}. 
The performance gain confirms that our 
problem
also benefits from these practices. 
Some qualitative comparisons are shown in~\cref{fig:synthetic-ablation,fig:detailed}.
Our ablation results and corresponding analysis show how \dataname helps benchmark model performance.

\qaheading{Does the view prompt \textcolor{GreenColor}{[$d$]} help the reconstruction} Yes. This is a common practice in SDS-based works~\cite{huang2024tech,chen2023fantasia3d,liao2023tada,poole2022dreamfusion}, yet has not been quantitatively justified.
As detailed in~\cref{table:ablation} (B. w/o view prompt), the normal error increased by \bbad{+9.3}. Apart from view prompts captioned by LLM, there is still room to improve with better camera representations, such as the camera pose embedding used in LGM~\cite{tang2024lgm} and ``Cameras-as-Rays''~\cite{zhang2024raydiffusion}.

\qaheading{Does NSFD outperform vanilla SDS} Yes. For a fair comparison, we set the guidance scale $s=7.5$ for both NSFD and vanilla SDS. As detailed in~\cref{table:ablation} (B. w/o NFSD), compared with NFSD (Noise-Free Score Distillation~\cite{katzir2024noisefree}), vanilla SDS degrades the geometry quality a bit by \bbad{+2.2}, while considerably degrading the texture quality (PSNR \bbad{+17.3}, LPIPS \bbad{+16.4}), as the SDS often crashes, leading to full-gray/yellow textures.

\begin{figure}[ht]
\caption{\textbf{Effectiveness of Synthetic Priors.} All the numbers refer to the performance gain (\%), where \textbf{\textcolor{greenprior}{Full}} means training with color-normal pairs, and \textbf{\textcolor{redprior}{RGB}} and \textbf{\textcolor{blueprior}{Normal}} means training with a single modality.}
\resizebox{0.9\linewidth}{!}{
\begin{tabular}{ccc}
  \includegraphics[trim=000mm 009mm 000mm 000mm, clip=true]{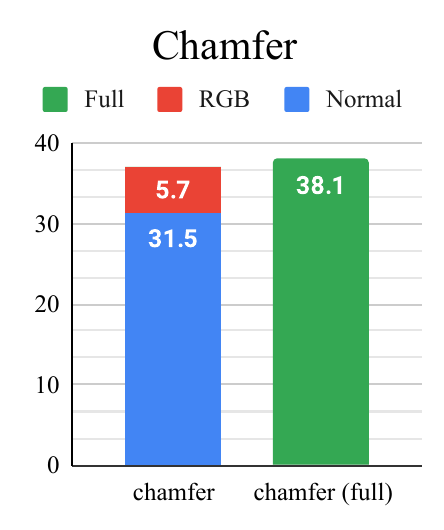} &   \includegraphics[trim=000mm 009mm 000mm 000mm, clip=true]{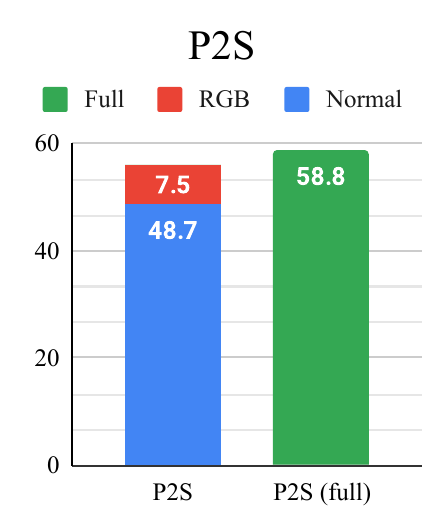} & \includegraphics[trim=000mm 009mm 000mm 000mm, clip=true]{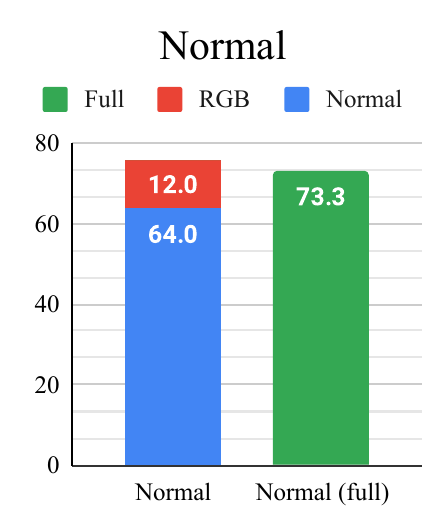} \\
 \includegraphics[trim=000mm 009mm 000mm 000mm, clip=true]{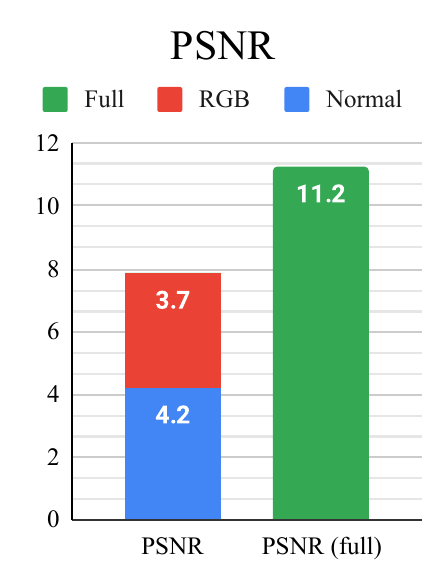} &   \includegraphics[trim=000mm 009mm 000mm 000mm, clip=true]{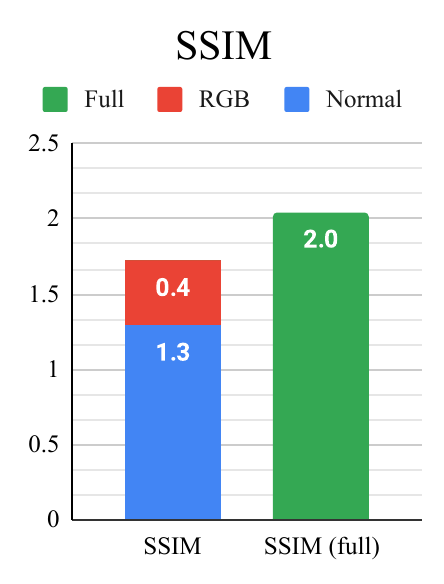} & \includegraphics[trim=000mm 009mm 000mm 000mm, clip=true]{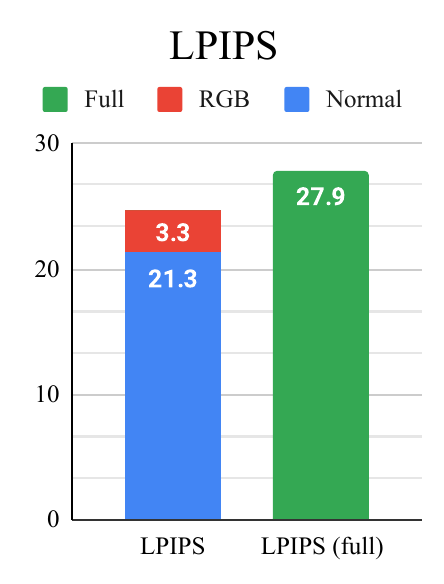}\\
\end{tabular}}
\label{fig:prior}
\end{figure}

\begin{figure*}[!htbp]
  \includegraphics[width=\linewidth]{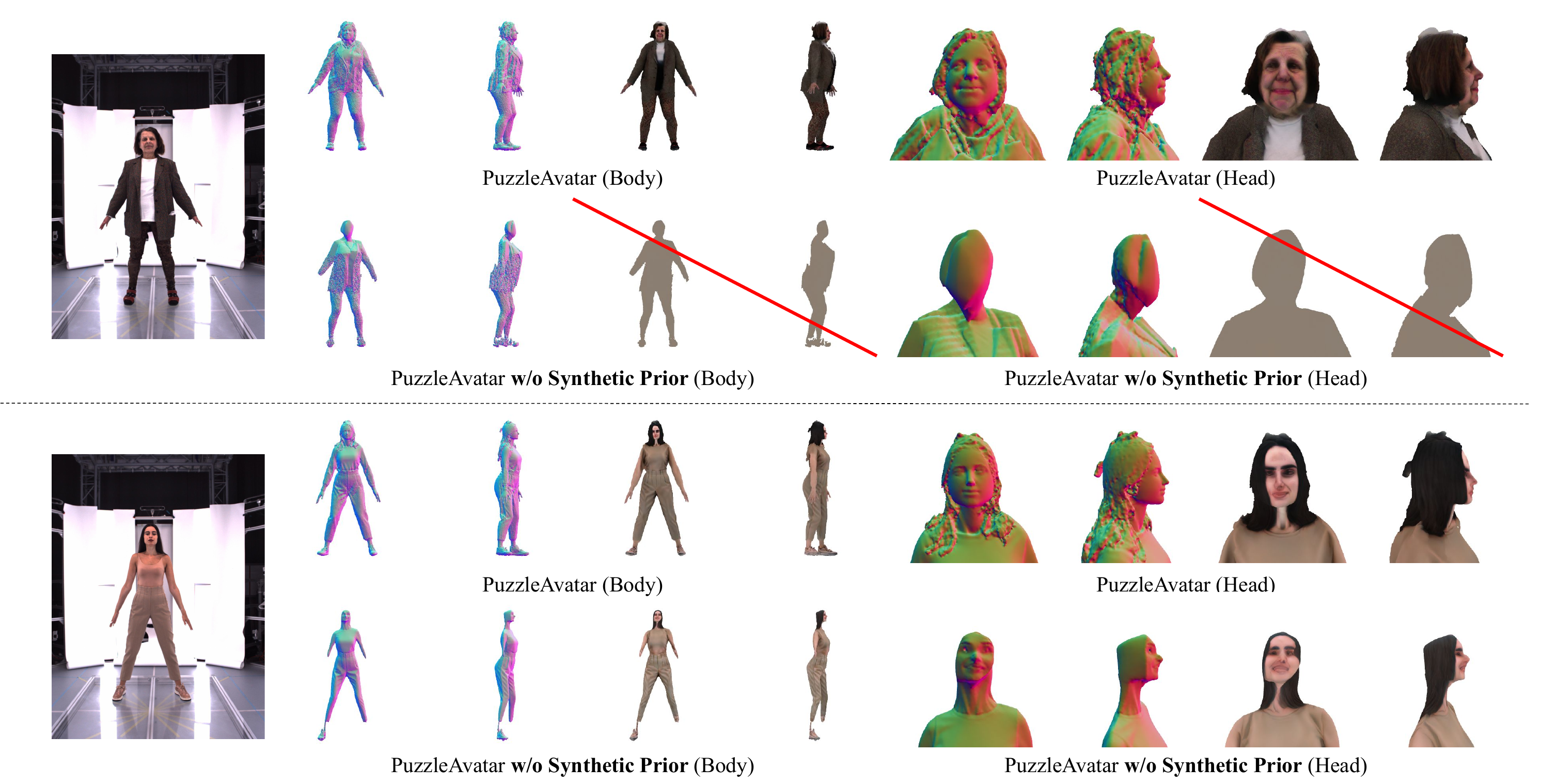}
  \vspace{-2.0 em}
  \caption{\textbf{How does Synthetic Prior Help?} See~
  \cref{fig:prior} for more in-depth analysis.}
  \label{fig:synthetic-ablation}
\end{figure*}

\qaheading{Does the synthetic human prior helps} 
Yes, and it significantly improves the reconstruction quality, in both the geometry (chamfer error \bgood{-38.1}, P2S error \bgood{-58.8}, Normal error \bgood{-73.3}), and texture (PSNR \bgood{+11.2}, LPIPS \bgood{-27.9}). And synthetic normals appear to contribute more than synthetic RGB (chamfer error \bgood{-31.5} vs. \bgood{-5.7}, LPIPS \bgood{-21.3} vs. \bgood{-3.3}). Introducing photorealistic synthetic data during fine-tuning proves beneficial, and the performance boost from color-normal pairs surpasses that from only using single mode (color/normal) of data, such as chamfer (\bbad{+38.1} > \bbad{+31.5} + \bbad{+5.7}) and LPIPS (\bbad{+27.9} > \bbad{+21.3} + \bbad{+3.3}), see~\cref{fig:prior}. We attribute the enhanced geometry-texture alignment to the pairwise training. Please check out~\cref{fig:synthetic-ablation} for more qualitative ablation results.

\begin{figure*}[!htbp]
  \includegraphics[width=\linewidth]{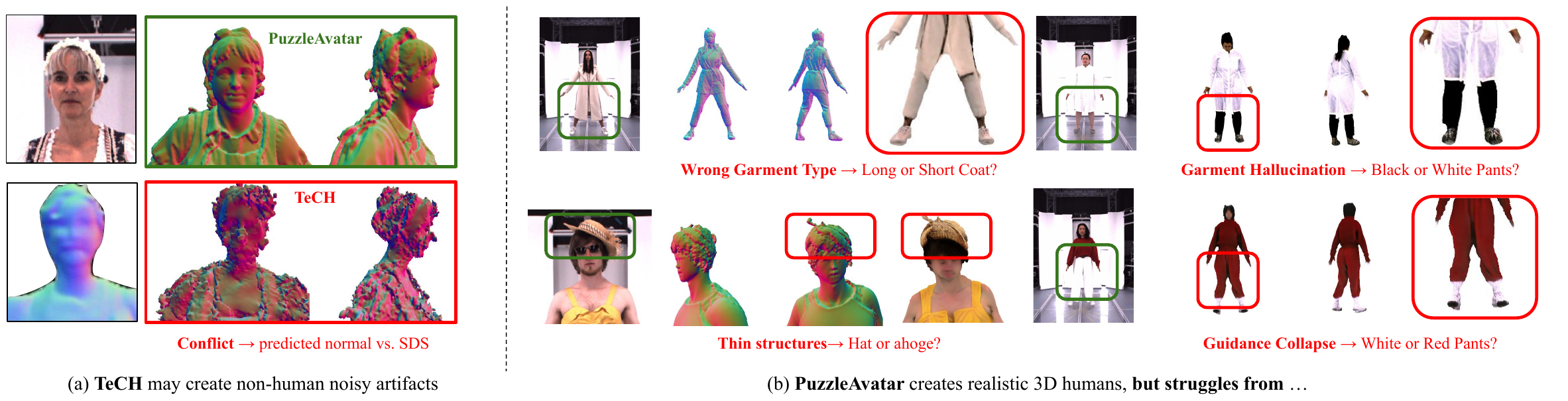}
  \caption{\textbf{Failure Cases.} Non-human artifacts are a common error for TeCH (see left part), whereas errors in \modelname stem from hallucination and flawed DMTet modeling of thin structures. For the right-top case, the black pants showing through the white coat, while realistic, deviate from the original input. As a result of this hallucination, the failures of \modelname cause differences from the ground-truth, but not completely catastrophic (see right part).}
  \label{fig:failure}
\end{figure*}

\qaheading{%
Can 
token \textcolor{blue}{[$v_i$]} %
\UPDATE{encode} 
the identity and features of assets}
Yes. 
As shown in~\cref{table:ablation} \mbox{(A. w/ detailed GPT-4V description)}, both shape and color quality slightly decrease when 
too-detailed
descriptions 
are used
\UPDATE{in} 
the prompt, such as ``\specific{wearing \underline{sleeveless} \textcolor{blue}{<asset1>} t-shirts, and \underline{fitted} \textcolor{blue}{<asset2>} jeans}'', 
instead of ``\specific{wearing \textcolor{blue}{<asset1>} t-shirts, and \textcolor{blue}{<asset2>} jeans}''. 
Surprisingly, more detailed prompts can introduce bias, conflicting with the original identity and harming performance; see \cref{fig:detailed}. \camera{It is worth mentioning that the same prompts are used for both training and testing to eliminate description mismatches. We attribute the identity mismatch to several potential reasons: (1) vocabulary mismatch between different text encoders (CLIP vs. GPT-4o), 
(2) guidance collapse, as the additional textual description of one garment may affect the appearance of others  (\cref{fig:failure}).}

\qaheading{Does \modelname %
work without using any full-body shots}
Yes but with some performance drop. 
Excluding the full-body shots, %
slightly decreases the quality of both geometry and texture (Chamfer \bbad{+0.9} and PSNR \bbad{-4.1}; ~\cref{table:ablation}, C. w/o full-body images). Nevertheless, 
it is unsurprising to find that 
\modelname without training on full-body images still outperforms the best TeCH setting (better texture plus on-par geometry quality).

\qaheading{How much data does \modelname need}
With just a 
\UPDATE{fraction of the training data}
(10\%), %
\modelname can already achieve satisfactory reconstruction performance.
As the number of training images increases (from 10\% to 50\%, in \cref{table:ablation}), \camera{the quality of view synthesis keeps improving, but not the geometry.}
We hypothesize that training \mbox{PuzzleBooth} using more RGB images could impair the qualiy of SDS gradients in the space of normal maps, thus degrading the geometry optimized via SDS. 
We find some empirical evidence supporting this hypothesis \cref{table:ablation} (B. without synthetic normal), where the absence of normal priors leads to a notable decline in geometry quality compared to texture (P2S \bbad{+48.7} vs. SSIM \bbad{-1.3}).

\qaheading{Does \modelname support multi-subject training}
Yes. 
\UPDATE{In fact, and perhaps surprisingly,} multi-subject training even slightly improves reconstruction quality (\cref{table:ablation}-C).
This demonstrates the power of Stable Diffusion to process and integrate numerous human identities simultaneously, and the robustness of our puzzle-based training strategy 
in learning disentangled human identities.

\section{Applications}

\begin{figure}
    \centering
    \includegraphics[width=\linewidth]{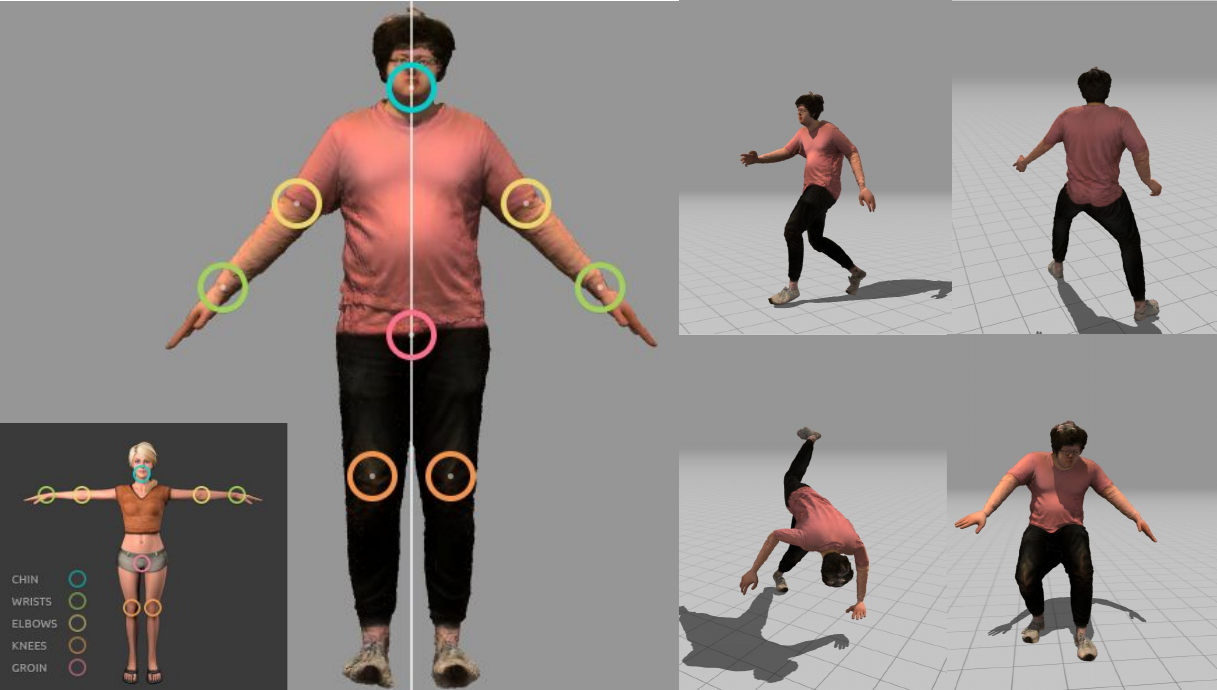}
    \caption{\camera{\textbf{Animatable PuzzleAvatar}. As the output of \modelname is A-posed, it could be easily rigged in ``symmetry-mode'' and then animated.}}
    \label{fig:pa-animation}
\end{figure}

The compositionality of \modelname through \UPDATE{its tokens and} text prompts supports diverse applications like Virtual Try-On and text-guided \UPDATE{avatar} editing, as shown in \CHECK{~\cref{fig:teaser}} and \vid.
\camera{Virtual Try-On, we fine-tuned a single diffusion model on two human-specific photo collections (A and B), where the learned tokens, initially repeated across both humans (A: \specific{<asset 1> face, <asset 2> shirt}; B: \specific{<asset 1> face, <asset 2> coat}), are uniquely assigned (A: \specific{<asset 1> face, <asset 2> shirt}; B: \specific{<asset 3> face, <asset 4> coat}). We describe this approach as ``multi-subject training'' in~\cref{sec:ablation}. During the virtual try-on process, we interchange tokens (\eg, \specific{<asset 1> face} $\Rightarrow$ \specific{<asset 3> face}).
The ability to do text-guided editing demonstrates that, PuzzleBooth could be learned without forgetting: as SD model is only fine-tuned on ``shirt'', yet it still produces a realistic ``coat'' when we replace ``\specific{<asset3> shirt}'' with ``\specific{<asset3> coat}''. In other words, the fine-tuned SD still remembers the concept of ``coat''.}
Moreover, the A-Posed output can simplify the rigging and skinning process, \camera{as shown with the Mixamo~\footnote{\url{https://www.mixamo.com/}} example at~\cref{fig:pa-animation}}. Besides, with the underlying \mbox{SMPL-X} parametric body, the 3D output could be easily animated with \mbox{SMPL-X} motion data, like \mbox{AMASS}~\cite{AMASS:ICCV:2019} and \mbox{AIST++}~\cite{li2021ai}, as is a the common practice in ~\cite{xiu2022icon, zheng2021pamir, Huang:ARCH:2020}. 

\section{Conclusion}

\lzheading{Limitations \& Future Work}
Since \modelname builds on PuzzleBooth and \sds, 
while using no 
re-projection terms, some hallucination is inevitable. As \cref{fig:failure} shows, \modelname may incorrectly hallucinate garment texture or types, and suffer from description contamination, a common issue in T2I models. Despite being trained with synthetic paired data, our model sometimes struggles to perfectly disentangle shape and color, causing geometric details like wrinkles to appear on textures. 
\camera{Regarding the geometry, \dmtet is not suitable for modeling very thin structures (\cref{fig:failure}-b), whiule spike-like noisy artifacts (\cref{fig:failure}-a) are inevitable when sculpting \dmtet using an SDS loss, a phenomenon also found in~\cite{huang2024tech,chen2023fantasia3d,gao2023contex}. Exploring new free-form 3D representations~\cite{gshell,son2024dmesh,guo2024tetsphere,Huang2DGS2024} and mesh-based deformers~\cite{aigerman2022neural,aanaes2016large,palfinger2022continuous} is worth trying to replace \dmtet. And a full 3D compositionality could be achieved through multi-layer 3D representations~\cite{Feng2022scarf,kim2024gala,wang2023disentangled,Zielonka2023Drivable3D}.}

\begin{figure}
    \centering
    \includegraphics[width=\linewidth]{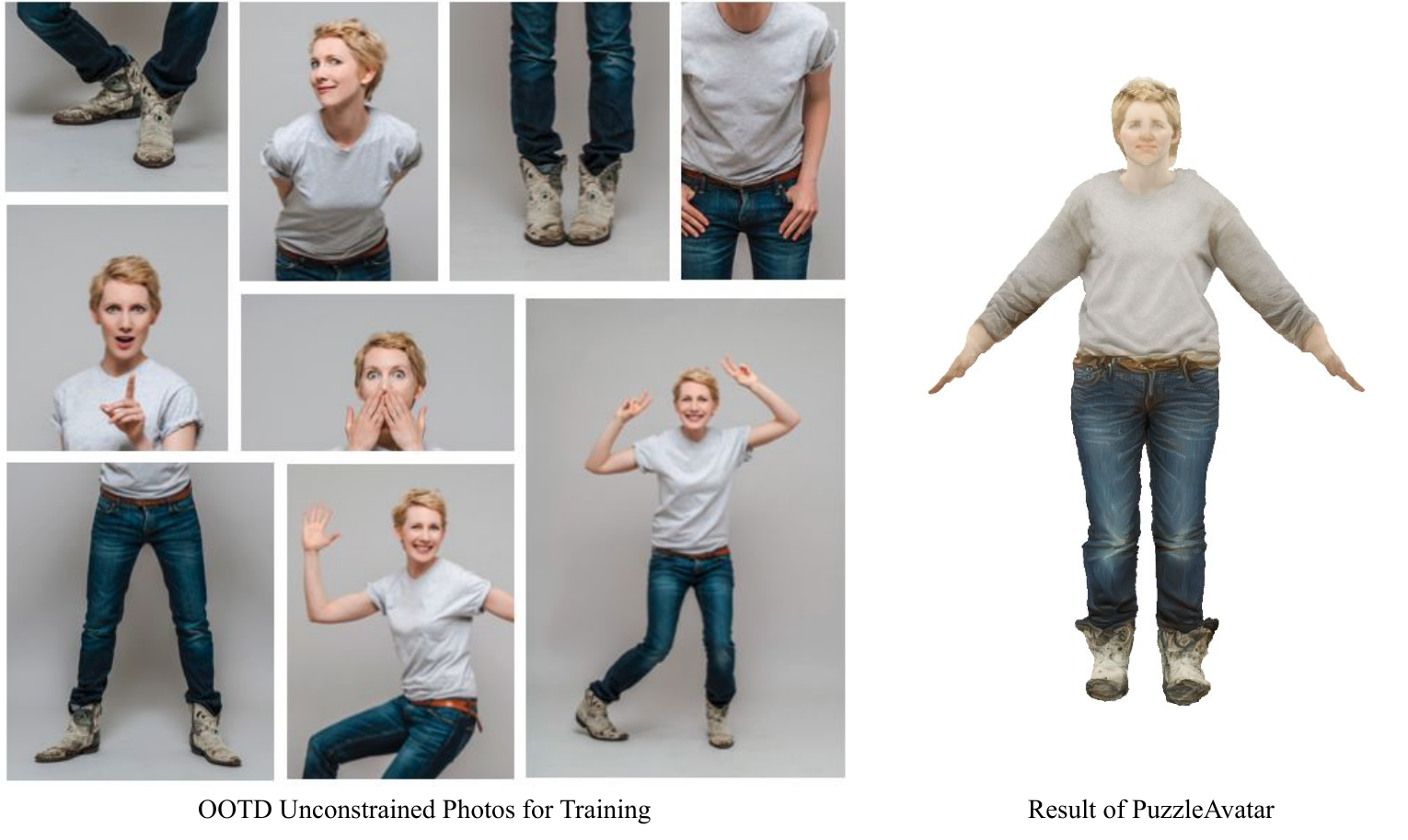}
    \caption{\camera{\textbf{Shape Inconsistency}. The person reconstructed by PuzzleAvatar is obviously much ``fatter'' than the real body.}}
    \label{fig:shape}
\end{figure}

Preserving facial identity is challenging without high-resolution headshots in the training data. Potential solutions for better identity preservation may include enhancing segmented faces with super-resolution techniques~\cite{basicsr}, conducting personalized restoration~\cite{chari2023personalized}, or incorporating face ID embeddings~\cite{Zielonka2022TowardsMR,wang2024instantid,feng2021pixie}. \camera{Additionally, shape inconsistency occasionally occur, as shown in~\cref{fig:shape}. A potential solution is to use a more accurate shape estimator, such as SHAPY~\cite{Shapy:CVPR:2022} and Semantify~\cite{Gralnik_2023_ICCV}, to derive a more precise body shape for initialization.}

\modelname's main issue currently is its computational complexity, %
as spending roughly 4 hours to train PuzzleBooth and perform SDS-based optimization is impractical for \camera{online or real-time reconstruction, and costly for offline training as well}. %
In the future we will explore 
better training-free strategies~\cite{li2023photomaker,tewel2024training,rout2024rb} and better sampling methods for diffusion models~\cite{luo2023lcm,song2023consistency}. 

Multi-subject training with \modelname seems promising. %
Thus, it might be feasible to extend \modelname to decentralized training settings.
By fine-tuning a shared T2I model through federated learning~\cite{jiang2023testtime}, users 
across the globe 
could upload their personal albums 
to build a global ``style set'' of really diverse 
clothing, 
accessories, 
and hairstyles, 
for customizing avatars.

\begin{figure*}[!htbp]
    \includegraphics[width=\linewidth]{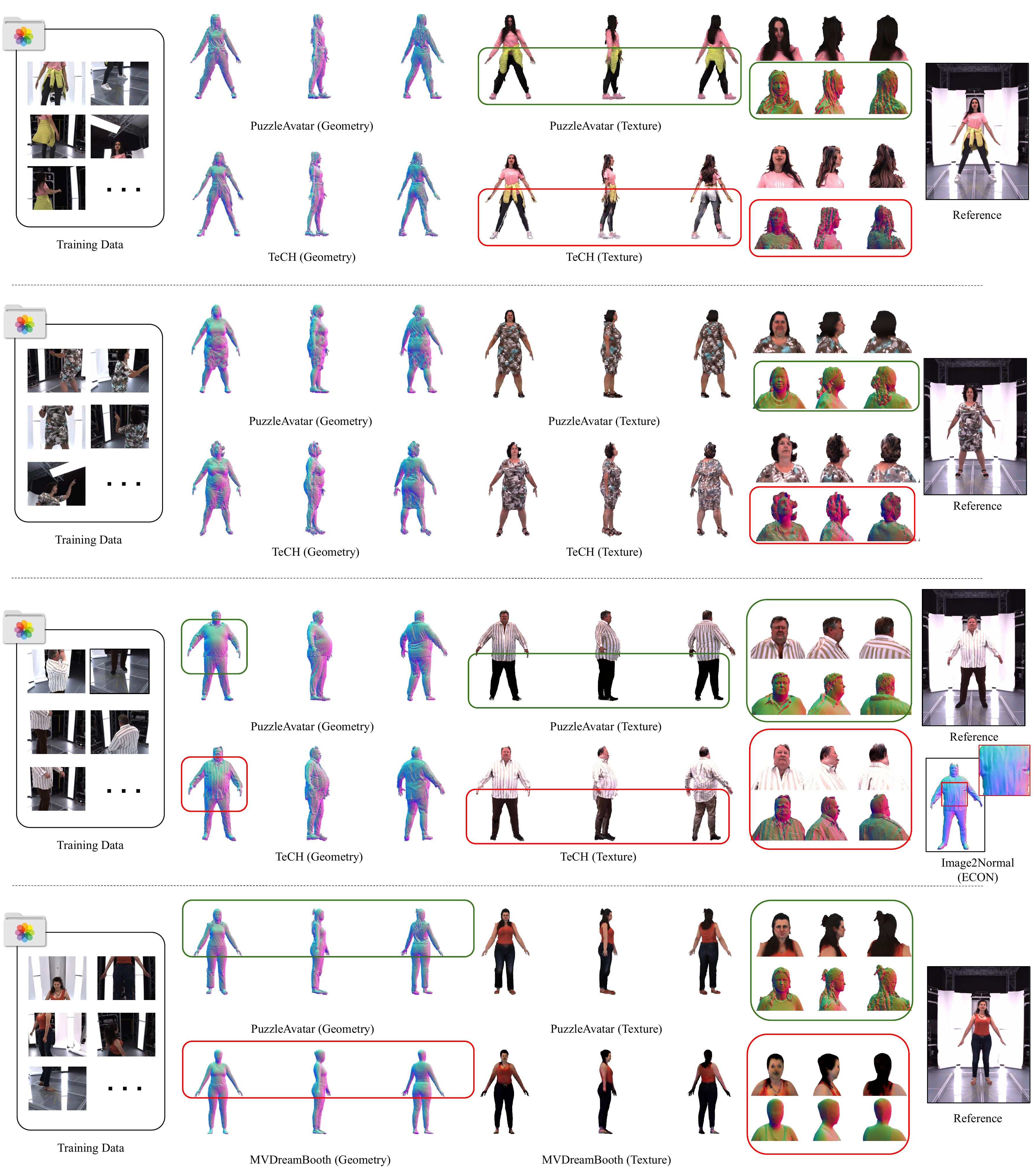}
    \caption{
    \textbf{\CHECK{Qualitative Results.}} 
        We compare \modelname, TeCH and \mbox{MVDreamBooth} on randomly sampled subjects. 
        \modelname offers various advantages over TeCH: 
        (1) Enhanced front-back consistency. 
        (2) Reduced non-human artifacts. 
        (3) Improved geometry-texture disentanglement. 
        At the bottom, the comparison with \mbox{MVDreamBooth} shows \modelname's capability in producing intricate details (shape+color).
        \faSearch~\textbf{Zoom in} to see more 3D and color details.
    }
    \label{fig:qualitative}
\end{figure*}

\camera{\lzheading{Indirect vs. Direct Reconstruction}
We acknowledge that our ``SDS-based person-specific generation loss'' (\textit{indirect reconstruction}) are less sensitive to fine-grained geometric misalignment (\eg, specific wrinkles in clothing, state of hair, or facial expression) than traditional ``re-projection loss'' (\textit{direct reconstruction}). \modelname leans more towards semantic-aligned rather than pixel-aligned reconstruction. This explains why the front-side rendering of TeCH always looks more pixel-aligned with the original input than \modelname, as shown in~\cref{fig:qualitative}. 
However, ``re-projection loss'' necessitate precise estimating camera, body pose or geometric maps (\ie, depth, normal), which is challenging in our unconstrained setting where both the human and camera move freely against random backgrounds. Thus, the pixel-aligned scheme is not scalable enough in the case of diverse unstructured inputs.
Finally, incorrect estimates of \textit{direct reconstruction} cause non-human artifacts in TeCH (\cref{fig:failure}-a), whereas errors of \modelname (\textit{indirect reconstruction}) mainly stem from model hallucination, as shown in \cref{fig:failure}-b, where the reconstructed shapes look realistic but vary slightly in identity.
%
}

\lzheading{Potential Negative Effect} 
As discussed in Sec.~\ref{sec:ablation}, the performance of \modelname relies heavily on existing public/commercial synthetic datasets and therefore may inherit their gender, racial and age biases. One may address such an issue by curating balanced datasets from real-world images (with off-the-shelf methods to 
estimate normals~\cite{xiu2022icon,xiu2022econ,saito2020pifuhd,bae2024dsine}) 
or by ``simply'' building better synthetic datasets.

\lzheading{Contributions to the Community}
\modelname paves the way for reconstructing detailed articulated humans from personal, natural photo collections -- introducing the new ``Album2Human'' task. Meanwhile, \dataname offers a new benchmark that facilitates \textit{objective} evaluation for various tasks, including but not limited to model customization, model personalization and distillation sampling. 
We believe that our new task, Album2Human, together with our new benchmark, \dataname, could push the boundary of the field of AI-Generated Content (AIGC). 
Furthermore, \modelname offers a simple yet scalable reconstruction system, with which users may %
ignore the technical details of reconstruction parameters. 
More importantly, we believe that \modelname demonstrates a new and practical paradigm for \textit{``puzzle-assembled clothed human reconstruction''} that produces a 3D avatar from everyday photos in a more scalable and constraint-free manner than \textit{``pixel-aligned clothed human reconstruction''}~\cite{saito2019pifu}. 

\appendix

\renewcommand{\thefigure}{R.\arabic{figure}}
\renewcommand{\thetable}{R.\arabic{table}}
\renewcommand{\theequation}{S.\arabic{equation}}

\setcounter{figure}{0}
\setcounter{table}{0}
\setcounter{equation}{0}


\clearpage

\begin{subappendices}
\renewcommand{\thesection}{\Alph{section}}%
\label{sec:appendix}

\section{GPT-4V Prompt for PuzzleBooth}
\label{sec:prompt-sup}

\qheading{Queried Prompt}
\textit{``Analyze the provided images, each featuring an individual. Identify and describe the individual's gender, facial features (excluding hair), haircut, and specific clothing items such as shirts, hats, pants, shoes, dresses, skirts, scarves, etc. Return the results in a dictionary format with keys for "gender", "face", "haircut", and each type of clothing. The corresponding value should provide 1-3 adjective or noun words, which describe the topological or geometric features, such as length (\eg, short, long, midi, mini, knee-length, floor length, ankle-length, hip-length, calf-length), shape (\eg, oval, round, square, heart-shaped, diamond-shaped, rectangular, voluminous, razor-cut, tousled, layered, messy), tightness (\eg, tight, snug, fitted, skin-tight, loose, tight-fitting, clingy), style (\eg, modern, casual, sporty, classic, formal, vintage, bohemian, avant-garde), or haircut types (\eg, long, short, wavy, straight, curly, bald, medium-length, pony tail, bun, plaits, beard, sideburns, dreadlocks, goatee), without referencing color or texture pattern. Exclude accessories and don't include any clothing item in the description of another. Omit any keys for which the clothing item does not appear or the description is empty. The response should be a dictionary only, without any additional sentences, explanations, or markdowns syntax (like json)''}

\camera{\section{Negative Prompt for SDS Optimization}
\label{sec:negative-prompt}
\qheading{Negative Prompt}
\textit{``unrealistic, blurry, low quality, out of focus, ugly, low contrast, dull, dark, low-resolution, gloomy, shadow, worst quality, jpeg artifacts, poorly drawn, dehydrated, noisy, poorly drawn, bad proportions, bad anatomy, bad lighting, bad composition, bad framing, fused fingers, noisy, many people, duplicate characters''}}

\section{Camera setting}
\label{sec:camera-sup}

To familiarize the diffusion model with the camera positions sampled during SDS optimization, we rendered the synthetic color-normal image pairs in the exact same manner as the SDS sampling strategy. This rendered data is used in preserving the synthetic human prior ($\mathcal{L}_\text{prior}$), while training the 2D generator $G_{\text{puzzle}}$. 

To ensure complete coverage of the entire body and face, we sample virtual camera poses around the full body and zoom in on the face region.
To reduce the occurrence of mirrored appearance artifacts (e.g., Janus-head), we incorporate view-aware prompts (\ie, ``\specific{front/side/back/overhead view}''), that describes the viewing angle during the generation process. The effectiveness of this approach has been demonstrated in DreamFusion~\cite{poole2022dreamfusion}.

To ensure full coverage of the entire body and the human face, we sample virtual camera poses into two groups: (1) $\mathbf{K}_\mathrm{body}$ cameras with a field of view (FOV) covering the full body or the main body parts, and (2) zoom-in cameras $\mathbf{K}_\mathrm{face}$ focusing the face region.

The ratio $\mathcal{P}_\mathrm{body}$ determines the probability of sampling $\mathbf{k}\in \mathbf{K}_\mathrm{body}$, while the height $h_\mathrm{body}$, radius $r_\mathrm{body}$, elevation angle $\phi_\mathrm{body}$, and azimuth ranges $\theta_\mathrm{body}$ are adjusted relative to the \smplx body scale.
Empirically, we set $\mathcal{P}_\mathrm{body}=0.5$, $h_\mathrm{body}=[-0.4, 0.4]$, $r_\mathrm{body}=(0.7, 1.3)$, $\theta_\mathrm{body}=[60^\circ, 120^\circ]$, $\phi_\mathrm{body}=[0^\circ, 360^\circ]$, with the $M_\mathrm{body}$ proportionally scaled to a $[-0.5, 0.5]$ unit space. 

To enhance facial details, we sample additional virtual cameras positioned around the face $\mathbf{k}\in \mathbf{K}_\mathrm{face}$, together with the additional prompt  ``\specific{face of}''. 
With a probability of $\mathcal{P}_\mathrm{face} = 1-\mathcal{P}_\mathrm{body} = 0.5$, the sampling parameters include the view target $c_\mathrm{face}$, radius range $r_\mathrm{face}$, rotation range $\theta_\mathrm{face}$, and azimuth range $\phi_\mathrm{face}$.
Empirically, we set $c_\mathrm{face}$ to the 3D position of SMPL-X head keypoint, $r_\mathrm{face}=[0.3, 0.4]$, $\theta_\mathrm{face}=[90^\circ, 90^\circ]$ and $\phi_\mathrm{face}=[-90^\circ, 90^\circ]$.

Regarding the synthetic data, we use all the subjects (525 textured scans) in THuman2.0. For each subject, we render 8 full-body views and 8 head views, as shown in~\cref{fig:synthetic-data}, and query their descriptive prompts via GPT-4V~\cite{gpt4v}. This gives us $525 \times 8 \times 2 = 8400$ color-normal pairs in total.

\end{subappendices}

\bigskip
\noindent\rule[3.0ex]{\linewidth}{1.5pt}

\qheading{Acknowledgments} 
We thank \textit{Peter Kulits} and \textit{Yandong Wen} for proofreading, \textit{Yifei Zeng} for providing the results of AvatarBooth, \textit{Yamei Chen} and \textit{Kexin Wang} for teaser photos, \textit{Jiaxiang Tang, Yangyi Huang, Nikos Athanasiou, Yao Feng} and \textit{Weiyang Liu} for fruitful discussions, \textit{Jinlong Yang} and \textit{Tsvetelina Alexiadis} for data capture. This project has received funding from the European Union’s Horizon $2020$ research and innovation programme under the Marie Skłodowska-Curie grant agreement No.$860768$ (\href{https://www.clipe-itn.eu}{CLIPE} project). \textit{Yufei Ye}'s PhD research is partially supported by a Google Gift.

\medskip

\qheading{Disclosure}
MJB has received research gift funds from Adobe, Intel, Nvidia, Meta/Facebook, and Amazon.  MJB has financial interests in Amazon and Meshcapade GmbH.  While MJB is a co-founder and Chief Scientist at Meshcapade, his research in this project was performed solely at, and funded solely by, the Max Planck Society.

\clearpage
\bibliographystyle{ACM-Reference-Format}
\bibliography{reference}

\end{document}